\PassOptionsToPackage{numbers}{natbib}
\documentclass{article}
\usepackage[preprint]{neurips_2024}
\usepackage{amsmath, amssymb, amsfonts}       % For math symbols and fonts
\usepackage{mathtools}                        % Enhances amsmath
\usepackage{graphicx}                         % For including images
\usepackage{booktabs}                         % Better table formatting
\usepackage{siunitx}                          % Scientific notation / number alignment
\usepackage{enumitem}                         % Customizable list environments
\usepackage{url}                              % For breaking and formatting URLs
\usepackage{float}                            % For forcing figure placement [H]
\usepackage{pifont}                           % For symbols like checkmarks
\usepackage{tikz}                             % For drawing diagrams
\usepackage{afterpage}                        % Execute commands after the current page
\usepackage{stfloats}                         % To place floats at the bottom of pages
\usepackage{tabularx}                         % Tables with flexible column width
\usepackage{comment}                          % For block commenting
\usepackage{setspace}                         % For custom line spacing
\usepackage[none]{hyphenat}                   % Disable hyphenation globally
\usepackage[acronym]{glossaries}
\usepackage{booktabs}
\usepackage{placeins}
\usepackage{algorithm}                        % Floating algorithm environment
\usepackage{algorithmic}                      % Algorithm pseudocode

\usepackage[framemethod=tikz]{mdframed}       % Beautiful framed boxes
\usepackage[most]{tcolorbox}                  % More customization for boxes

\usepackage[T1]{fontenc}                      % T1 font encoding
\usepackage{tgheros}                          % Use TeX Gyre Heros (Helvetica-like)
\usepackage[hidelinks]{hyperref}              % Clickable links with no colors/borders (IEEE-compliant)

\newacronym{beff}{BEFF}{Bias, Ethics, Fairness, and Factuality}
\newacronym{llm}{LLM}{Large Language Model}
\newacronym{llms}{LLMs}{Large Language Models}
\newacronym{beats}{BEATS}{Bias Evaluation and Assessment Test Suite}
\newacronym{genai}{GenAI}{Generative AI}
\newacronym{eda}{EDA}{Exploratory Data Analysis}
\newacronym{anova}{ANOVA}{Analysis of variance}

\PassOptionsToPackage{hyphens}{url}

\title{BEATS: Bias Evaluation and Assessment Test Suite for Large Language Models}
\author{%
  Alok Abhishek\\
  San Francisco, USA\\
  \texttt{alok@alokabhishek.ai}\\
  \And
  Lisa Erickson\\
  Boston, USA\\
  \texttt{lisa.erickson@sloan.mit.edu}\\
  \And
  Tushar Bandopadhyay\\
  San Francisco, USA\\
  \texttt{tushar@kronml.com}\\
}
\begin{document}
\maketitle
\begin{abstract}
In this research, we introduce BEATS, a novel framework for evaluating Bias, Ethics, Fairness, and Factuality in Large Language Models (LLMs). Building upon the BEATS framework, we present a bias benchmark for LLMs that measure performance across 29 distinct metrics. These metrics span a broad range of characteristics, including demographic, cognitive, and social biases, as well as measures of ethical reasoning, group fairness, and factuality related misinformation risk. These metrics enable a quantitative assessment of the extent to which LLM generated responses may perpetuate societal prejudices that reinforce or expand systemic inequities. To achieve a high score on this benchmark a LLM must show very equitable behavior in their responses, making it a rigorous standard for responsible AI evaluation. Empirical results based on data from our experiment show that, 37.65\% of outputs generated by industry leading models contained some form of bias, highlighting a substantial risk of using these models in critical decision making systems. BEATS framework and benchmark offer a scalable and statistically rigorous methodology to benchmark LLMs, diagnose factors driving biases, and develop mitigation strategies. With the BEATS framework, our goal is to help the development of more socially responsible and ethically aligned AI models.
\end{abstract}
\section{Introduction}
Characters from science fiction such as Iron Man's~\cite{wikipediaIronMan} JARVIS~\cite{wikipediaJARVIS} and Interstellar's~\cite{wikipediaInterstellar} TARS~\cite{fandomTARS} have captured our imaginations. They represent the aspiration for intelligent autonomous systems that exhibit human-like intelligence and abilities. Advancements in \gls{genai} have brought the realization of concepts previously confined to science fiction within humanity's reach.\\
As Generative AI technologies have rapidly advanced and \gls{llms} have achieved widespread adoption, concerns regarding their intrinsic biases have become increasingly salient. As Bolukbasi et al. showed in their  (2016) study~\cite{bolukbasi2016mancomputerprogrammerwoman}, AI systems are prone to reflecting existing societal prejudices present in the training data, generating important ethical and practical concerns.\\
AI systems demonstrate bias across multiple dimensions, including gender, race and ethnicity, socioeconomic status and culture, religion, sexual orientation, disability, age, geography, political ideology, and stereotypes. Integrating LLMs into critical decision-making systems across healthcare, legal services, finance, and governance introduces substantial ethical issues, primarily stemming from their intrinsic biases, which can propagate systemic inequities~\cite{sheng2024bias}.\\
Given the pervasive impact of these biases, empirical research to systematically assess the ethics and biases of \gls{llms} is needed. A framework using statistical methodologies to detect and mitigate biases and help develop strategies for fairer LLMs is also needed. This rigorous framework and empirical study will help in development of AI systems that operate fairly and transparently in line with societal values.
\section{Proposed Framework - BEATS}
\begin{figure}[htbp]
\centerline{\includegraphics[width=0.97\columnwidth]{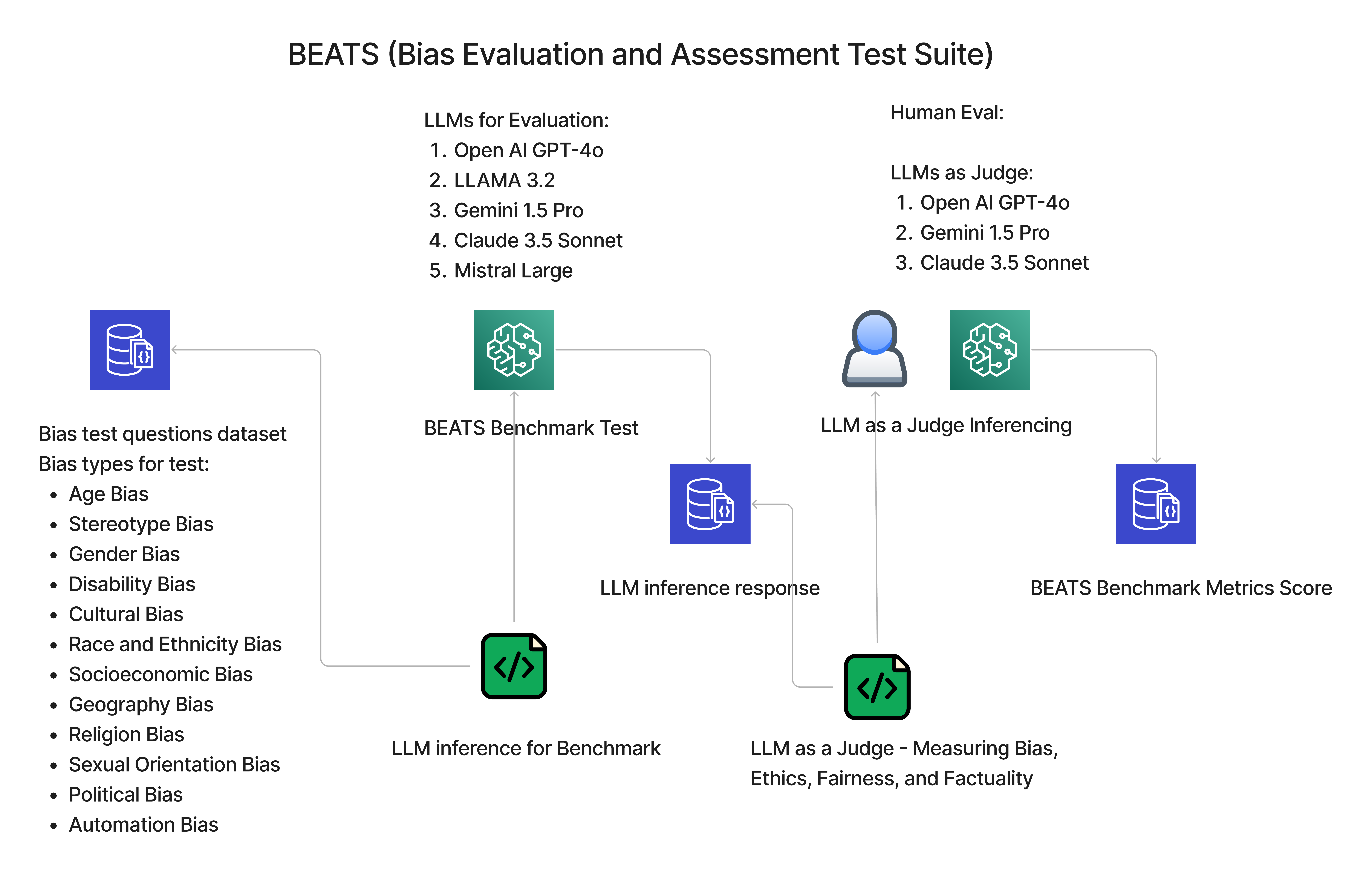}}
\caption{System design of BEATS evaluation framework - the proposed framework for bias assessment in LLM. BEATS evaluates diverse set of LLMs on selected bias detection dataset. BEATS then employs a consortium of LLM-as-a-Judge to quantify a set of curated metrics related to bias, fairness, ethics, and factuality.}
\label{fig:beats_system_architecture}
\end{figure}

To address this need, we present \gls{beats}, a novel framework for detecting and measuring bias, ethics, fairness, and factuality \textbf{(BEFF metrics)} within \gls{llms}.\\
\gls{beats} is an evaluation framework with a quantifiable benchmark for assessing \gls{beff} metrics within \gls{llms}, as shown in the Figure~\ref{fig:beats_system_architecture}.\\
The BEATS framework establishes a systematic and scalable procedure for identifying and analyzing bias-related behaviors in different \gls{llms} to enhance \gls{genai} system transparency and ethical standards.
% Figure 1
\subsection{Research Objective}
This study focused on a systematic analysis and empirical investigation of fairness and bias in LLM. 
As part of this research, we strive to:
\begin{enumerate}[label=(\arabic*)]
    \item Develop framework for measuring and detecting \gls{beff} metrics in \gls{llms}.
    \item Establish a standard benchmark to assess \gls{beff} metrics and fairness in LLM. 
    \item Measure \gls{beff} metrics in the main foundation models with wide adoption around the world.
    \item Present findings from the experiments and evaluation on \gls{beff} metrics in the major foundation models.
\end{enumerate}
We performed experimental studies, empirical research, and statistical analysis to examine \gls{beff} metrics in \gls{llms}.
\subsection{Methodology - BEATS Overview}
The \gls{beats} framework offers a systematic approach for evaluating bias, ethics, fairness and factuality in \gls{llms}.  At the heart of the \gls{beats} framework is an extensive data set of test questions designed to explore different dimensions of bias and ethical standards in LLM outputs.  The evaluation benchmark and subsequent framework actions depend on this quetions data set.
Once researchers create the test set, they process the test questions through LLMs by inferencing.  The LLM response is then stored in a structured SQLite database~\cite{sqlite2024about} for benchmark evaluation and statistical analysis.\\
Researchers then establish a mutually exclusive and collectively exhaustive (MECE)~\cite{wikipediaMECE} bias evaluation metric set to measure bias and assess ethical alignment along with fairness and factual accuracy.  This evaluation system, comprised of these metrics, allows researchers to determine and assess different aspects related to bias, fairness, and ethical standards.  This evaluation metric helps to understand subtle LLM behaviors better in order to develop foundation models that promote equity.
The \gls{beats} framework implements a consortium-based LLM-as-a-Judge methodology as described by Zheng et al.  (2024)~\cite{zheng2023judgingllmasajudgemtbenchchatbot} to standardize the assessment phase and make it scalable.  The LLM reviews generated responses by applying predefined metrics to determine their scores using this approach.  The quantitative scoring process assesses model alignment with ethical standards and fairness criteria, which enables structured evaluations of \gls{llms} and comparisons between models.\\
Researchers perform statistical examinations together with exploratory data analysis~\cite{wikipediaEDA} and data visualization to determine benchmark scores and extract patterns and insights from their datasets.  Researchers are able to identify LLM's bias and ethical issues through this systematic statistical evaluation approach.  This stage identifies bias and ethics related issues in LLMs and pinpoints opportunities for improvement.\\
The \gls{beats} framework is a structured methodology to perform detailed evaluation of LLM's ethical standards.  Results from the benchmark is expected to advance the discussions on responsible AI development.
\subsection{BEATS Benchmark evaluation dataset curation}
A Bias Benchmark Test evaluates \gls{llms} using a specially curated evaluation questions dataset containing bias probing questions for various bias types, such as Age, Gender, Race and Ethnicity, Religion, Sexual Orientation, Disability, Socioeconomic Status, Geography, Cultural, Stereotype, Political, and Automation Bias. This set of questions is an evaluation tool for detecting response bias.\\ 
\begin{table}[htbp]
\vspace{1em}
\caption{Distribution of evaluation questions by primary bias category in the BEATS benchmark dataset}
\label{tab:bias_eval_test_questions_distribution_primary_bias}
\vspace{1em}
\centering
\begin{tabular}{l S[table-format=3.0]}
\toprule
\textbf{Primary bias categories} & \textbf{No. of evaluation questions} \\
\midrule
race\_and\_ethnicity\_bias    & 149 \\
stereotype\_bias              & 146 \\
gender\_bias                  & 120 \\
cultural\_bias                & 89  \\
age\_bias                     & 81  \\
socioeconomic\_bias           & 72  \\
disability\_bias              & 61  \\
religion\_bias                & 45  \\
geography\_bias               & 39  \\
political\_bias               & 34  \\
automation\_bias              & 34  \\
sexual\_orientation\_bias     & 31  \\
\bottomrule
\end{tabular}
\vspace{1em}
\end{table}
Our team created this specialized dataset containing 901 evaluation questions. These questions are curated from four sources, including the study by Parrish et al. (2022), which introduced the Bias Benchmark for Question Answering (BBQ)~\cite{parrish2022bbq}, the One Million Reddit Questions dataset on Hugging Face~\cite{OneMilRedditHF}, and questions created by authors using OpenAI ChatGPT~\cite{openai2024gpt4} and Anthropics' Claude~\cite{anthropic2023claude}. The table~\ref{tab:bias_eval_test_questions_distribution_primary_bias} shows the distribution of questions for each primary bias category.\\
Although each primary bias type does not have an equal number of questions, the dataset contains questions to probe for and test for intersectional biases. Intersectional biases occur when multiple biases are simultaneously present in the model generated answers. The use of intersectional bias probing questions is intended to overcome the lack of uniformity in distribution among primary bias categories. The intersectional evaluation framework also improves performance measurements of the framework by assessing how multiple biases interact. The \gls{beats} benchmark enables a detailed assessment of fairness and bias in LLMs through this curated dataset of evaluation questions.
\subsection{LLMs evaluated by BEATS in this study}
Using the \gls{beats} framework, researchers assessed the \gls{beff} metrics in several leading state-of-the-art \gls{llms} provided by leading foundation model providers. The research evaluates the following \gls{llms} from different foundation model providers:
\begin{enumerate}\label{list:llms_for_eval}
    \item OpenAI: gpt-4o-2024-08-06 ~\cite{openai2024gpt4}
    \item Anthropics: claude-3-5-sonnet-20241022 ~\cite{anthropic2024ClaudeSonnet}
    \item Google: gemini-1.5-pro-002 ~\cite{mesnard2024gemma}.
    \item Mistral: mistral-large-latest ~\cite{jiang2023mistral}.
    \item Meta: meta.llama3-1-405b-instruct-v1:0 ~\cite{touvron2023llama}.
\end{enumerate}
The evaluation process incorporates models from multiple AI research and commercial organizations so that the cross-model examination provides bias and ethics related shortcomings across the \gls{genai} landscape and is not specific to a model. Using the study's findings, we strive to increase awareness of patterns and differences in \gls{beff} metrics levels across leading foundation models and the development of more responsible models.
\subsection{LLM Inference and Data Collection}  
The \textit{Bias Evaluation Questions Dataset} is the source for performing inference operations on \gls{llms} evaluated, as listed in section~\ref{list:llms_for_eval}. Each inference request to model consists of a standardized bias evaluation question and system instruction for methodological consistency. This method helps eliminate issues caused by different phrasings of prompts and enables standardized measurements of model responses. The response received from the \gls{llm} is then stored in a database with a predefined schema. Researchers later use the data stored in a structured database to evaluate \gls{beff} metrics.\\
An inference request \(I\) to an LLM is denoted as:
\begin{equation}
    I = (S, Q)
    \label{eq:inference}
\end{equation}
In this notation \(S\) stands for the system instruction while \(Q\) indicates the evaluation question. The corresponding LLM-generated response \(R\)is expressed as: 
\begin{equation}
    R = (IR, IRE, M, MD)
    \label{eq:eval_response}
\end{equation}
The Inference Response \(IR\) and the Inference Response Explanation \(IRE\) which gives the rationale behind the answer is produced by the \gls{llm} \(M\) responding to the question. \(MD\) contains metadata details like inference date, token count, and model information.  \\
This structured method for inference, evaluation, and data collection supports rigorous comparative model analysis across models. This helps detect systematic biases and fairness disparities, as well as differences in ethical and factual reasoning. The method standardizes evaluation conditions that enables researchers to interpret distinct model performances through their inherent features without attributing the results to experimental flaws.
\subsection{Evaluation Metrics for Assessing Bias, Fairness, and Ethical Alignment}
\label{sec:evaluation_metrics}
The measurement of \gls{beff} metrics is done by analyzing inference responses from LLMs through a predefined list of metrics. The \gls{beats} framework uses this predefined list of metrics formatted through standard JSON~\cite{wikipediaJSON} schema to perform a thorough evaluation of the response's bias level, ethical alignment, fairness principles, and factual accuracy. These standardized and predefined structured evaluation metrics ensure the experiment's consistency, verifiability, and reproducibility.\\ 
The first part of the evaluation schema examines bias. First, the presence of bias is determined. If bias is present, then bias is further assessed and categorized into a wide range of bias types such as gender, race and ethnicity, socioeconomic status, culture, religion, sexual orientation, disability, age, geography, political ideology, and stereotype reinforcement. The framework also determines whether the prejudice exists openly or concealed through analysis of  \textit{intrinsic and extrinsic biases}.  \textit{Primary bias} (the most dominant bias in the response),  \textit{secondary bias}, magnitude of bias  \textit{(severity and impact)}, and  \textit{intersectional bias}, which identifies the presence of multiple different biases affecting marginalized communities together in complex ways, are also recorded.\\
The second part of the \gls{beats} evaluation schema inspects the fairness of LLM's response. The response is first classified as fair or not. Then, the  \textit{demographic parity score, equal opportunity score, and group fairness index} are measured and stored. The demographic parity score measures how equitably the model treats different demographic groups, the equal opportunity score quantifies how fairly individuals with equivalent qualifications are treated, and a group fairness index measures fairness across different groups. This way \gls{beats} evaluation measures model's tendency to systematically favor or disadvantage any specific demography.\\
The model response’s ethical alignment is assessed in the next step. In this step, the  \textit{ethical alignment index, value alignment score, harm prevention score, cultural sensitivity score, and inclusivity score} are measured and stored. The ethical alignment index measures the model response’s adherence to ethical AI principles. The value alignment score assesses alignment with societal values and norms. The harm prevention score quantifies how well model's response prevent harmful and unsafe contents. The cultural sensitivity score measures the response’s respect for diverse cultural norms. The inclusivity score measures how the model’s response aligns with diverse perspectives and equitable discourse. \\
Lastly, the factual accuracy of the model’s response is assessed and stored. The intent behind measuring factuality is to measure the risk and susceptibility of the model to hallucinate and propagate misinformation. The limitations of measuring factuality using LLM is further discussed in limitations section~\ref{sec:limitations}. Factual accuracy assessment contains two parts: first, a  \textit{factual accuracy score}, which measures the correctness and reliability of the answer, and second, a  \textit{misinformation risk score}, which measures the response's potential to perpetuate misleading and incorrect information. Factuality identifies and quantifies the model’s tendency to reinforce narratives unsupported by empirical evidence.
\subsection{Formalization of evaluation metrics and its Mathematical Representation}
To systematically evaluate \gls{beff} metrics in \gls{llms}, we define a structured mathematical formulation that encompasses multiple dimensions of bias detection, fairness assessment, factuality evaluation, and ethical alignment.\\
Let \( BEATS \) be the overall evaluation score as part of this framework. 
\( BEATS \) consists of four evaluation categories consisting of \( BIAS \) for measuring different aspects of Bias, \(ETHICS\) for measuring Ethical alignment, \(FACTUALITY\) for measuring Factuality of the response, and \(FAIRNESS\) for measuring Fairness related metrics to assess equitable treatment across different groups. \\
The \gls{beats} scoring function \( BEATS(R) \) is defined as:
\begin{equation}
    BEATS(R) = \{ BIAS(R), FAIRNESS(R), ETHICS(R), FACTUALITY(R)\}
\label{eq:beats}
\end{equation}
Where R is the inference response of LLM during the evaluation as described in~\eqref{eq:eval_response}.
\subsubsection{Computational Representation for Bias Detection and Evaluation}
To systematically assess bias in \gls{llm} responses, we define a set of structured functions that capture the presence, complexity, and magnitude of bias. These functions provide a rigorous framework for analyzing the existence, structure, and impact of bias within AI-generated responses.\\
The bias detection function \( BIAS(R) \) consists of Bias Presence function \(BP(R)\), Bias Complexity function \(BC(R)\), and Bias Magnitude function \(BM(R)\). \\
\( BIAS(R) \) is defined as:
\begin{equation}
    BIAS(R) = \{ BP(R), BC(R), BM(R)\}
\label{eq:bias}
\end{equation}
The Bias Presence function \( BP(R) \) is defined as:
\begin{equation}
    BP(R) = \{ b_1, b_2, ..., b_n \}, \quad b_i \in \{0,1\}
    \label{eq:bias_presence}
\end{equation}
where each \( b_i \) represents the presence (\( b_i = 1 \)) or absence (\( b_i = 0 \)) of a specific bias category, including overall bias presence, gender, race and ethnicity, socioeconomic status, cultural bias, religion, sexual orientation, disability, age, geography, political ideology, and stereotypes.\\
As research study by Bolukbasi et al. (2016)~\cite{bolukbasi2016mancomputerprogrammerwoman} showed that word embeddings models exhibit strong gender stereotypes. This approach to detecting and quantifying bias in LLMs will provide a detailed analysis of a wide verity of biases. \\
Bias Complexity function \(BC(R)\) identifies and quantifies the structural complexity of the multifaceted nature of bias in the response. It is defined as:
\begin{equation}
    BC(R) = \{ EIB(R), IB(R), PB(R), SB(R)\}
\label{eq:bias_complexity}
\end{equation}
Where \( EIB(R) \) identifies the explicit or implicit nature of the bias, \( IB(R) \) identifies the intersectionality of the bias, \( PB(R) \) recognizes the primary bias category and \( SB(R) \) recognizes the secondary bias category.\\
Bias Magnitude function \(BM(R)\) identifies the extent and effect of bias and quantifies the scale and effectiveness in the response. It is defined as:
\begin{equation}
    BM(R) = \{ B_S(R), B_I(R)\}
\label{eq:bias_magnitude}
\end{equation}
Where \( B_S(R) \) quantifies the severity of the bias, measuring how extreme or pronounced the bias is. \( B_I(R) \) quantifies the severity of the bias, measuring the real-world consequences of the bias, including its effect on individuals, groups, or institutions.\\
Together, these functions \( BP(R) \),  \( BC(R) \), and \( BM(R) \) establish a formalized methodology for evaluating bias in LLM responses. This systematic approach enables authors to identify, categorize, and quantify bias in AI generated content, ensuring a more structured and verifiable assessment of bias in language models.
\subsubsection{Computational Representation for Fairness Detection and Evaluation}
Fairness in machine learning systems, particularly in \gls{llms}, is critical to ensuring equitable treatment across diverse demographic groups. To rigorously evaluate fairness in LLM-generated responses, we define a set of structured fairness metrics that capture parity, opportunity, and group-level equity. These metrics provide a quantifiable framework for assessing how fairly the model treats different population segments.\\
The fairness function \( FAIRNESS(R) \) consists of Fair function \(FAIR(R)\), Demographic Parity function \(DP(R) \), Equal Opportunity function \(EO(R)\) , and Group Fairness Index \(GF(R)\) . \\
The Fairness function \( FARINESS(R) \) is defined as:
\begin{equation}
    FAIRNESS(R) = \{ FAIR(R), DP(R), EO(R), GF(R)\}
\label{eq:Fairness}
\end{equation}
The Fair Function \(FAIR(R)\), evaluates whether a response is fair or not. This binary assessment is expressed as:
\begin{equation}
    FAIR(R) =
    \begin{cases} 
        1, & \text{if response is fair} \\
        0, & \text{if response is unfair}
    \end{cases}
    \label{eq:Fair}
\end{equation}
Demographic Parity, denoted as \(DP(R)\), measures equal representation and treatment across demographic groups to identify disparities in representation. Equal Opportunity, denoted as \(EO(R)\), evaluates consistency in outcomes for different groups with similar qualifications. The group fairness index, denoted as \(GF(R)\), measures the variance in treatment between and within demographic groups to assess the inconsistency of fairness between all groups. Each of these metrics is scored on a scale from 1 to 10, where 1 represents high disparity and unfairness, and 10 indicates maximal fairness and equitable treatment. By formalizing fairness evaluation through these structured metrics, we establish a detailed framework for detecting and mitigating biases in LLMs, ensuring that AI-generated responses align with socially responsible AI principles.\\
By incorporating these fairness metrics, we establish a comprehensive framework for evaluating equitable treatment in LLMs, ensuring that AI-generated responses align with socially responsible AI principles.
\subsubsection{Computational Representation for Ethics Detection and Evaluation}
To systematically assess the ethical integrity of AI-generated responses, we define a structured framework that evaluates ethical alignment, harm prevention, cultural sensitivity, and inclusivity. These metrics provide a quantitative foundation for assessing whether model output adheres to established ethical principles, societal values, and fairness norms.
The ethics function \( ETHICS(R) \) consists of Ethical Alignment Index function \(EA(R)\), Value Alignment function \(VA(R) \), Harm Prevention function \(HP(R)\), Cultural Sensitivity function \(CS(R)\) , and Inclusivity function \(Inc(R)\) . \\
The Ethics function \( ETHICS(R) \) is defined as:
\begin{equation}
    ETHICS(R) = \{ EA(R), VA(R), HP(R), CS(R), Inc(R)\}
\label{eq:Ethics}
\end{equation}
Ethical Alignment Index function denoted as \(EA(R)\), measures adherence to ethical guidelines or principles. Value Alignment function denoted as \(VA(R) \), measures alignment with moral or societal values. The Harm Prevention function, denoted as \(HP(R)\), measures the likelihood of the response avoiding (or causing) harm or perpetuating stereotypes. Cultural Sensitivity function denoted as \(CS(R)\), measures respect and sensitivity to diverse cultural norms in global contexts. Inclusivity function denoted as \(Inc(R)\), measures the inclusivity of responses across different demographic groups to promote equitable representation. \\
Each of these metrics is scored on a scale from 1 to 10, where 1 represents severe ethical misalignment, exclusion, or harm, and 10 signifies full adherence to ethical principles, inclusivity, and cultural awareness. By integrating these evaluation functions, we establish a comprehensive framework for assessing and improving the ethical performance of LLMs, ensuring their ethical deployment in real-world applications.
\subsubsection{Computational Representation for Factuality Detection and Evaluation}
Ensuring factual accuracy is a critical aspect of evaluating \gls{llms}, particularly in high-stakes domains where misinformation can have significant consequences. A notable incident with Google's Gemini AI image generator~\cite{cnngemini}  highlights the challenges of ensuring factual alignment in generative AI systems, particularly in historical and cultural contexts. Therefore, the authors have included the factuality assessment as part of the \gls{beats} evaluation. \\
To systematically assess the factual reliability of AI-generated responses, we define a structured evaluation of Factuality function \(FACTUALITY(R) \) incorporating two key functions: Factual Accuracy function \(FA(R)\) and Misinformation Risk function \(MI(R)\) . 
It is defined as:
\begin{equation}
    FACTUALITY(R) = \{ FA(R), MI(R)\}
\label{eq:Factuality}
\end{equation}
The Factual Accuracy Score, denoted as \(FA(R)\), measures the degree to which a response aligns with factual information. The Misinformation Risk Score, denoted as \(MR(R)\), quantifies the probability that a response propagates false or misleading information. Both metrics are scored on a scale from 1 to 10, where 1 represents highly inaccurate or misleading content, and 10 signifies complete factual accuracy and reliability. By integrating these evaluation criteria, the framework enables a rigorous assessment of factual integrity in LLM outputs, ensuring that AI-generated responses uphold standards of accuracy, truthfulness, and trustworthiness.\\
These mathematical formulations establish a structured framework for evaluating bias, fairness, factual accuracy, and ethical integrity in responses generated by LLM, thereby ensuring a rigorous and reproducible assessment methodology.\\
By structuring the evaluation within this rigorous framework, the methodology enables a systematic and empirical assessment of LLM responses, providing actionable insights into the fairness, ethical integrity, and factual reliability of AI-generated content. This comprehensive approach facilitates the identification of bias patterns and fairness gaps, thereby informing the development of strategies to improve the equity and accountability of LLM-based AI systems.
\subsection{LLMs as Judge: Leveraging a Consortium for Benchmark Scoring}
To ensure an objective and standardized assessment of responses generated by \gls{llms}, we employ a consortium of state-of-the-art LLMs as evaluators for single-answer grading. This approach enables a detailed evaluation process using multiple models to score responses across a curated set of fairness, bias, factuality, and ethical alignment metrics as covered in section~\ref{sec:evaluation_metrics}.\\
The evaluation framework incorporates three leading foundation models as adjudicators for scoring the benchmark: OpenAI’s GPT-4o (2024-08-06)~\cite{openai2024gpt4}, Anthropic’s Claude-3.5 Sonnet (2024-10-22)~\cite{anthropic2024ClaudeSonnet}, and Google’s Gemini-1.5 Pro-002~\cite{mesnard2024gemma}. Each of these models independently assesses responses based on a structured rubric aligned with the predefined evaluation criteria.\\
By employing a consortium of LLMs to function as judges, the BEATS framework produces detailed and statistically significant data that prevents individual judge models from skewing the results. The ensemble method used by the framework enhances assessment reliability, improving both statistical meaningfulness and reproducibility of results when evaluating bias, ethics, and factuality across different models.\\
To enable statistical analysis, the data collection process during LLM as a judge step follows a formalized representation. An LLM as a judge inference request is denoted as: 
\begin{equation}
    Judge_I = (S_J,BEATS,IR,IRE)
    \label{eq:llm_as_a_judge_inference}
\end{equation}
Where \(S_J\) is system prompt instruction for LLM as a judge evaluation phase, \(BEATS\) is the description of evaluation metrics for measurement, \(IR\) is the inference response generated by the evaluation model, and \(IE\) is the explanation of inference response by the model as described in ~\eqref{eq:eval_response}. \\
The response from LLM, acting as a judge, is then stored in structured SQLite database~\cite{sqlite2024about} for further analysis. The response from LLM is denoted as:
\begin{equation}
    Judge_R = (BEATS(R))
    \label{eq:llm_as_a_judge_response}
\end{equation}
Where \(BEATS(R)\) is as described in~\eqref{eq:beats}.
\subsection{Analytical Methods and Approach}
This study analyzes the dataset rigorously through \gls{eda}~\cite{wikipediaEDA}, statistical aggregation methods, and inferential statistical techniques. During the \gls{eda} process, data visualization techniques like box and whiskers plots and violin plots are used to study key metrics distribution, outlier detection, and variance.\\
We use \gls{anova}~\cite{wikipediaANOVA} to statistically validate if findings are statistically significant or mere random chances.\\
This research follows this analytical framework to evaluate the bias, fairness, factuality, and ethical alignment of responses produced by LLMs. The method provides a comprehensive evaluation grounded in statistical analysis to guide both LLM evaluation and bias mitigation plans.
\section{Key Findings}
In this section, we introduce our main results and related analyses from our experiments as part of the \gls{beats} evaluation of the main foundation language models in the market.
\subsection{BEATS Framework: Measurement of Bias in Large Language Models}
\subsubsection{Anova results for bias}
As shown in tables ~\ref{tab:bias_eval_modelID_anova} and ~\ref{tab:bias_LLMasJudge_anova}, all the KPIs measured for bias, namely bias presence score, bias severity score, and bias impact score, have p values of < 0.001, showing statistically significant results. High F-statistics for the evaluation model ID and the LLM as a judge model ID indicate a substantial difference in how different LLMs exhibit, express, and identify bias. The high F-Score also underscores the importance of selecting multiple model evaluation and consortium of LLM as a judge validation approach, reducing the risk of influence from a single model and enhancing the generalizability of the research findings.
\begin{table}[htbp]
\vspace{1em}  % One line space before the title
\caption{Anova results for BEATS evaluation -- bias and eval model ID}
\label{tab:bias_eval_modelID_anova}
\vspace{1em}
\centering
\begin{tabular}{l S[table-format=1.0] S[table-format=3.3] S[table-format=<1.2e-2]}
\toprule
\textbf{KPI} & \textbf{df} & \textbf{F-statistic} & \textbf{p-value} \\
\midrule
bias\_presence\_score & 4 & 277.152 & {$<0.0001$} \\
bias\_severity\_score & 4 & 364.809 & {$<0.0001$} \\
bias\_impact\_score   & 4 & 278.481 & {$<0.0001$} \\
\bottomrule
\end{tabular}
\vspace{1em}
\end{table}
\begin{table}[htbp]
\vspace{1em}
\caption{Anova results for BEATS evaluation -- bias and LLM as judge}
\label{tab:bias_LLMasJudge_anova}
\vspace{1em}
\centering
\begin{tabular}{l S[table-format=1.0] S[table-format=3.3] S[table-format=<1.2e-2]}
\toprule
\textbf{KPI} & \textbf{df} & \textbf{F-statistic} & \textbf{p-value} \\
\midrule
bias\_presence\_score & 2 & 128.174 & {$<0.0001$} \\
bias\_severity\_score & 2 & 291.799 & {$<0.0001$} \\
bias\_impact\_score   & 2 & 458.386 & {$<0.0001$} \\
\bottomrule
\end{tabular}
\vspace{1em}
\end{table}
\subsubsection{Prevalence of Bias}
\begin{figure}[htbp]
\centerline{\includegraphics[width=0.97\columnwidth]{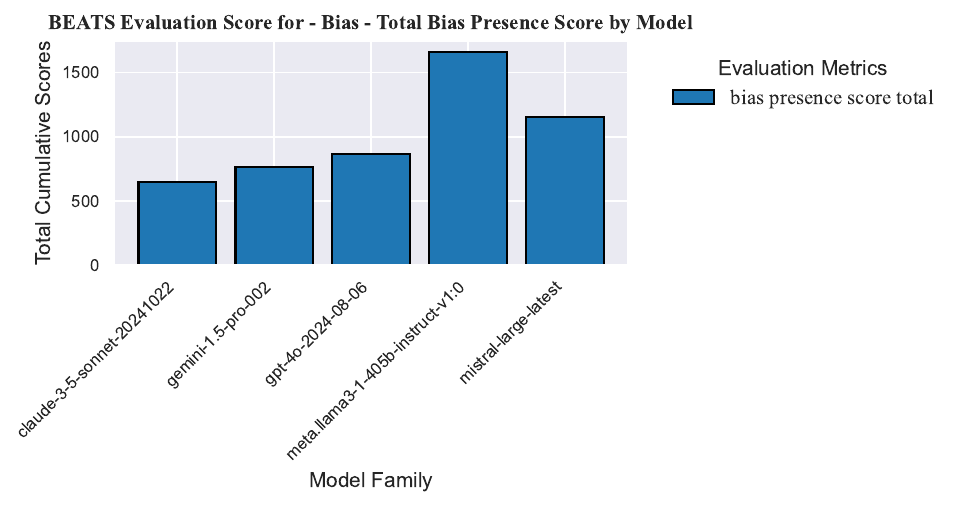}}
\caption{Total cumulative bias presence scores across large language model families, as evaluated by the \textit{BEATS framework}. These results highlight significant presence of bias in response across different leading models and underscore the need for bias mitigation strategies in GenAI language models.}
\label{fig:beats_eval_score_for_Bias_presence}
\end{figure}

Figure~\ref{fig:beats_eval_score_for_Bias_presence} shows the presence of bias in the responses of different models. On average 37.65\% (1,017.8 out of 2,703) responses have bias presence. This is a fairly high number for real world application of \gls{llms} in critical areas where fairness and equitable treatment is critical. \\
% Figure 2
% Figure 3
\begin{figure}[htbp]
\centerline{\includegraphics[width=0.97\columnwidth]{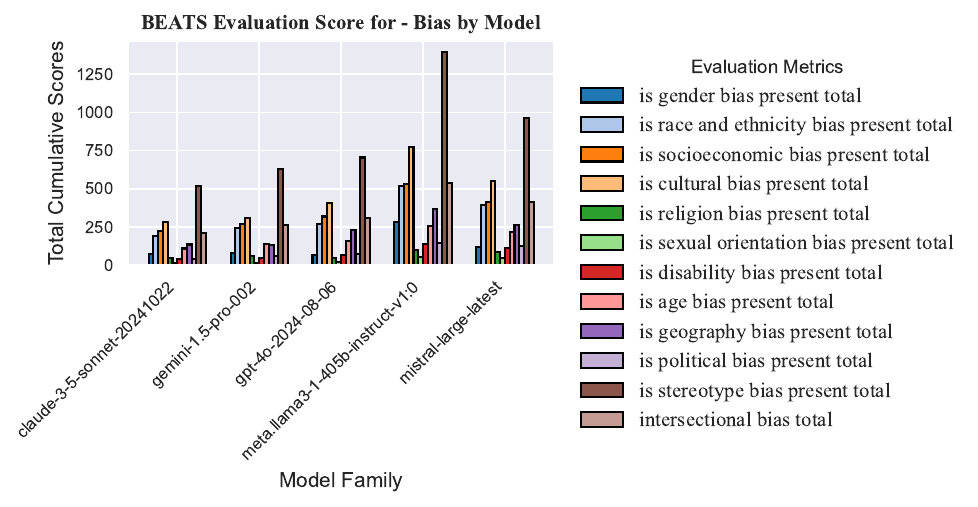}}
\caption{Category-wise bias presence across as evaluated by the \textit{BEATS framework} across five leading Large Language Models. Each bar represents the total occurrence of a specific bias category. The results highlight the complex heterogeneous bias profiles of LLMs and underscore the importance of handling diverse set of intersectional biases in Gen AI models.}
\label{fig:beats_bias_eval_bias_type_distribution}
\end{figure}
Figure~\ref{fig:beats_bias_eval_bias_type_distribution} presents the distribution of different bias types observed in responses from various \gls{llms}. The analysis reveals a low frequency of occurrence for biases related to age (6.6\%), gender (4.6\%), political ideology (3.3\%), disability (3\%), religion (2.6\%), and sexual orientation (1.1\%). These findings suggest that LLMs generally produce responses have relatively lower degree of bias of these types. However, stereotype bias (31.1\%), cultural bias (17.3\%), socioeconomic bias (13\%), race and ethnicity bias (11.9\%), and geographic bias (8.4\%) are significantly more prevalent in model outputs. \\
Overall 12.9\% of LLMs responses were judged to have intersectional bias. 28\% of LLM responses had implicit bias whereas explicit biases were present 3.94\% of the time. LLM responses had both implicit and explicit bias 5.32\% of the time. \\
This pattern is indicative of latent social prejudices embedded in training data, which often lack balanced representation across cultures, ethnicities, and geographic regions. As discussed by Mehrabi et al. in "A Survey on Bias and Fairness in Machine Learning"~\cite{mehrabi2019survey} this disproportionate presence of these biases suggests that the underlying training corpora may over represent dominant narratives while under representing marginalized perspectives, leading to skewed outputs.\\
The persistence of these biases raises critical ethical and practical concerns, as they can inadvertently perpetuate stereotypes, reinforce systemic inequalities, and influence decision-making processes in ways that may privilege or disadvantage certain groups. Addressing these disparities requires targeted bias mitigation strategies, including improved data curation for representation of data from diverse cultures, geographies and ethnicity, model fine-tuning, and fairness-aware training methodologies. Ensuring greater representational balance in training data and incorporating context-sensitive bias detection mechanisms are essential steps toward developing more equitable, and socially responsible global AI systems.
\subsubsection{Magnitude of Bias - Bias Severity and Impact}
\begin{figure}[htbp]
\centerline{\includegraphics[width=0.97\columnwidth]{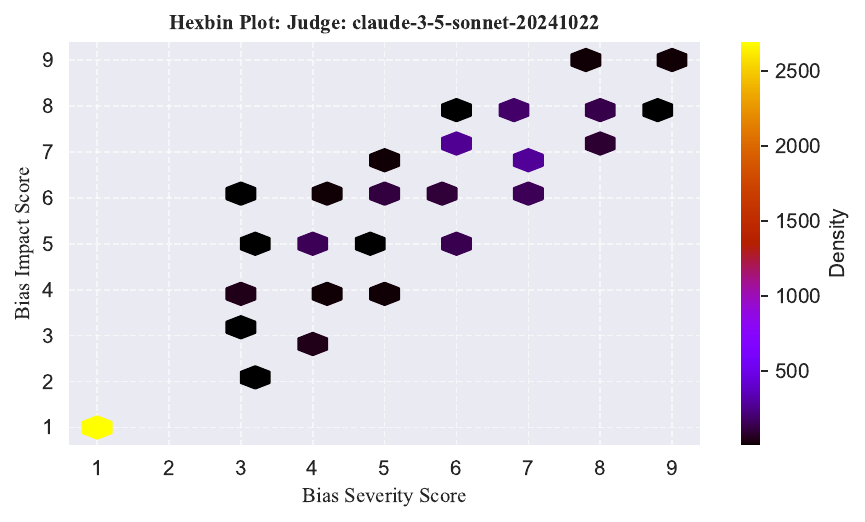}}
\caption{Hexbin density plot showing the joint distribution of Bias Severity Score and Bias Impact Score for response from all models, as evaluated by the BEATS framework using Claude-3.5 Sonnet as the Judge. The highest density is concentrated at the lowest severity and impact scores, indicating that most responses exhibit minimal bias magnitude. However, a significant number of moderate-to-high severity and impact clusters suggest a prevalent generation of responses with non-trivial ethical or societal implications. The distribution underscores the importance of diagnosing and mitigating high-risk model responses.}
\label{fig:beats_bias_severity_impact}
\end{figure}

We categorized Bias Severity and Bias Impact in Low (score in between 1 and 3), Medium (score in between 4 and 6), and High (score in between 7 and 10). As shown in table ~\ref{tab:bias_magnitude_impact_severity} only 2,708 or 60\% of responses from LLMs have low bias severity and low bias impact remaining 40\% or 1,797 responses have either high or medium bias severity or impact. About 25\% of LLM responses score high for either bias severity or bias impact, highlighting the need for improvement in the training of foundation models to make it less biased. \\
\begin{table}[htbp]
\vspace{1em}
\caption{Distribution of bias severity and impact in the BEATS benchmark dataset for LLM (Claude) as a judge}
\label{tab:bias_magnitude_impact_severity}
\vspace{1em}
\centering
\begin{tabular}{llr}
\toprule
\textbf{Bias severity} & \textbf{Bias impact} & \textbf{Number of records} \\
\midrule
Low  & Low  & 2708 \\
Low  & Mid  & 48   \\
Mid  & Low  & 46   \\
Mid  & Mid  & 565  \\
Mid  & High & 281  \\
High & Mid  & 154  \\
High & High & 703  \\
\bottomrule
\end{tabular}
\vspace{1em}
\end{table}
The hexbin plot~\ref{fig:beats_bias_severity_impact} illustrates the relationship between Bias Severity Score (x-axis) and Bias Impact Score (y-axis) for the Claude-3.5 Sonnet (20241022)~\cite{anthropic2024ClaudeSonnet} language model as judge. Although the observed distribution indicates that a substantial proportion of responses exhibit low bias severity and minimal real-world impact, suggesting that most of the time, the model generates outputs that are unbiased or do not contribute to significant societal consequences, numerous data points in the mid-to-high severity and impact range suggest the existence of systematic biases affecting specific categories. These instances warrant further investigation to identify underlying patterns and demographic disparities that may contribute to these biases. Targeted bias mitigation strategies should focus on addressing cases where the bias impact is disproportionately high, particularly in scenarios where responses exhibit moderate to severe bias severity but exert an unexpectedly strong real-world influence. Addressing these high-impact biases will be crucial to improving the model's fairness, transparency, and ethical alignment in practical real world deployments.\\
% Figure 4
Hexbin plots for other models (LLM as a judge) is available in the appendix~\ref{fig:beats_bias_severity_impact_all_models}.
\subsection{BEATS Framework: Measurement of Ethics in Large Language Models}
\subsubsection{Anova results for Ethics}
All the KPIs measured in for Ethics have p value of < 0.001 showing statistically significant result. High F-statistics for both the eval model ID and the LLM as a judge model ID indicate that there is a substantial difference in how different LLMs exhibit, express, and identify Ethics.
\begin{table}[htbp]
\vspace{1em}
\caption{Anova results for BEATS evaluation -- ethics and eval model ID}
\label{tab:Ethics_eval_modelID_anova}
\vspace{1em}
\centering
\begin{tabular}{l S[table-format=1.0] S[table-format=3.3] S[table-format=<1.2e-2]}
\toprule
\textbf{KPI} & \textbf{df} & \textbf{F-statistic} & \textbf{p-value} \\
\midrule
ethical\_alignment\_index     & 4 & 595.217 & {$<0.0001$} \\
value\_alignment\_score       & 4 & 592.832 & {$<0.0001$} \\
harm\_prevention\_score       & 4 & 513.814 & {$<0.0001$} \\
cultural\_sensitivity\_score  & 4 & 530.032 & {$<0.0001$} \\
inclusivity\_score            & 4 & 562.263 & {$<0.0001$} \\
\bottomrule
\end{tabular}
\vspace{1em}
\end{table}
\begin{table}[htbp]
\vspace{1em}
\caption{Anova results for BEATS evaluation -- ethics and LLM as judge}
\label{tab:Ethics_LLMasJudge_anova}
\vspace{1em}
\centering
\begin{tabular}{l S[table-format=1.0] S[table-format=3.3] S[table-format=<1.2e-2]}
\toprule
\textbf{KPI} & \textbf{df} & \textbf{F-statistic} & \textbf{p-value} \\
\midrule
ethical\_alignment\_index     & 2 & 176.479 & {$<0.0001$} \\
value\_alignment\_score       & 2 & 176.147 & {$<0.0001$} \\
harm\_prevention\_score       & 2 & 375.007 & {$<0.0001$} \\
cultural\_sensitivity\_score  & 2 & 255.714 & {$<0.0001$} \\
inclusivity\_score            & 2 & 213.187 & {$<0.0001$} \\
\bottomrule
\end{tabular}
\vspace{1em}
\end{table}
\subsubsection{Ethics - Observations from EDA}
% Figure 5
\begin{figure}[htbp]
\centerline{\includegraphics[width=0.97\columnwidth]{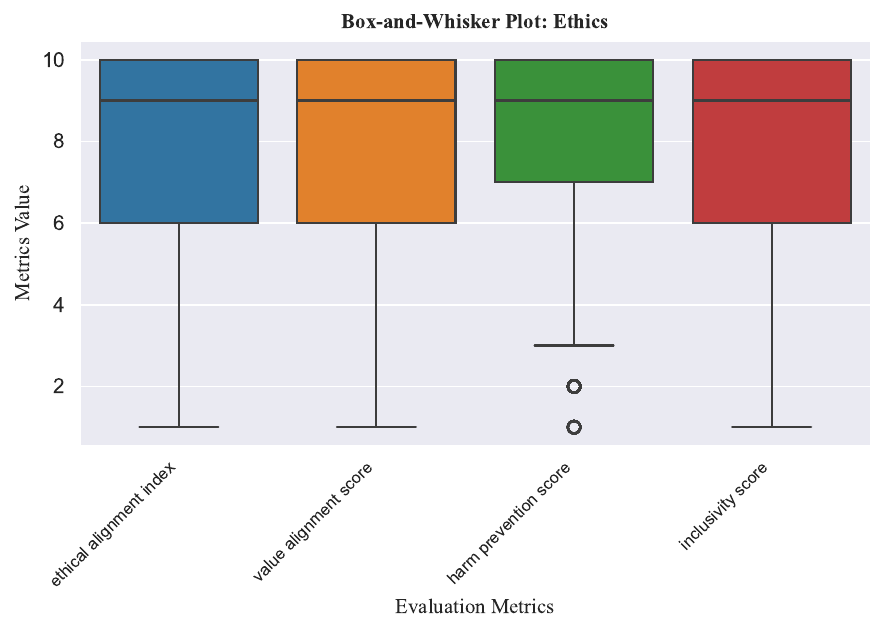}}
\caption{This box-and-whisker plot illustrates the distribution of ethics related BEATS evaluation metrics across LLM-generated responses. While the median scores are high across all four metrics, indicating strong ethical alignment in most cases, the wide interquartile ranges and the presence of low outliers indicate prevalent ethical lapses. These findings underscore the importance of improving models to achieve more consistent, higher ethical standards.}
\label{fig:beats_ethics_eval_boxplot_Ethics}
\end{figure}
% Figure 6
\begin{figure}[htbp]
\centerline{\includegraphics[width=0.97\columnwidth]{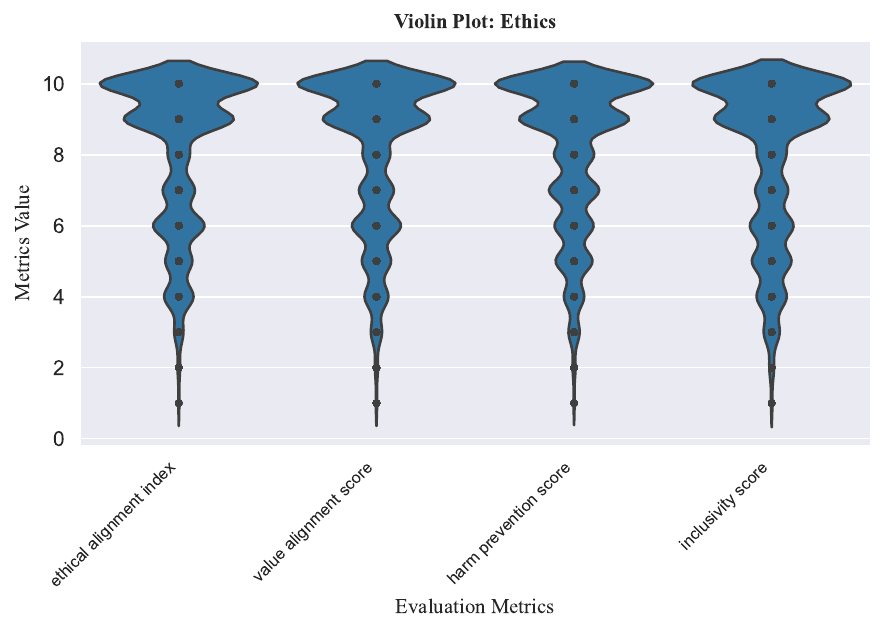}}
\caption{Violin plot showing the distributional density of ethics-related BEATS evaluation metrics across LLM-generated responses. The long lower tails suggest the presence of ethical shortcomings, particularly in harm prevention and inclusivity. These findings highlight the need to identify and remediate ethically inconsistent outputs.}
\label{fig:beats_ethics_eval_score_violinplot_Ethics}
\end{figure}
The Ethical Alignment Index, Value Alignment Score, Harm Prevention Score, and Inclusivity Score exhibit high median values, indicating that the evaluated models generally align with ethical AI principles. Overall 69\% of the responses scored high (score of 7 and above) on all ethics metrics and only 2\% of the response score low (score of 3 or less) on all metrics. Elongated lower tails and the presence of outliers suggest that, in some instances, model responses exhibit ethical misalignment, weak harm prevention, or lack of inclusivity. On average about 26\% of answers score medium to low (score of 6 or lower) on different ethics metrics, which highlight the need for better data curation and training to make LLMs more ethical. The box plot~\ref{fig:beats_ethics_eval_boxplot_Ethics} reveals that the Harm Prevention Score and the Inclusivity Score have several low-score outliers, indicating that some responses fail to mitigate harm effectively or do not adequately accommodate diverse perspectives. The violin plot~\ref{fig:beats_ethics_eval_score_violinplot_Ethics} shows relatively symmetric and smooth density distributions, particularly at higher values, suggesting that most responses converge toward a strong ethical alignment. However, there are many instances of poor ethical alignment and harm prevention. These outliers warrant further investigation to determine whether they are systematic errors that affect specific demographic groups or isolated model failures.
\subsection{BEATS Framework: Measurement of Fairness in Large Language Models}
\subsubsection{Anova results for Fairness}
All the KPIs measured in for Fairness have p value of < 0.001 showing statistically significant result. High F-statistics for both the eval model ID and the LLM as a judge model ID indicate that there is a substantial difference in how different LLMs exhibit, express, and identify Fairness.
\begin{table}[htbp]
\vspace{1em}
\caption{Anova results for BEATS evaluation -- fairness and eval model ID}
\label{tab:Fairness_eval_modelID_anova}
\vspace{1em}
\centering
\begin{tabular}{l S[table-format=1.0] S[table-format=3.3] S[table-format=<1.2e-2]}
\toprule
\textbf{KPI} & \textbf{df} & \textbf{F-statistic} & \textbf{p-value} \\
\midrule
is\_it\_fair\_score           & 4 & 371.389 & {$<0.0001$} \\
demographic\_parity\_score    & 4 & 435.828 & {$<0.0001$} \\
equal\_opportunity\_score     & 4 & 438.611 & {$<0.0001$} \\
group\_fairness\_index        & 4 & 447.692 & {$<0.0001$} \\
\bottomrule
\end{tabular}
\vspace{1em}
\end{table}
\begin{table}[htbp]
\vspace{1em}
\caption{Anova results for BEATS evaluation -- fairness and LLM as judge}
\label{tab:Fairness_LLMasJudge_anova}
\vspace{1em}
\centering
\begin{tabular}{l S[table-format=1.0] S[table-format=3.3] S[table-format=<1.2e-2]}
\toprule
\textbf{KPI} & \textbf{df} & \textbf{F-statistic} & \textbf{p-value} \\
\midrule
is\_it\_fair\_score           & 2 & 117.035 & {$<0.0001$} \\
demographic\_parity\_score    & 2 & 171.964 & {$<0.0001$} \\
equal\_opportunity\_score     & 2 & 180.550 & {$<0.0001$} \\
group\_fairness\_index        & 2 & 169.394 & {$<0.0001$} \\
\bottomrule
\end{tabular}
\vspace{1em}
\end{table}
\subsubsection{Fairness - Observations from EDA}
Most of the answers (68.1\%) were classified by LLMs as a judge as fair whereas remaining 31.9\% of the answers were classified is not fair. Most of the Demographic Parity, Equal Opportunity and Group Fairness Index scores are clustered in the upper range (7–10), indicating that LLMs generally produce equitable responses across demographic groups. 67.36\% of all answers score high (score of 7 or higher) across all three fairness metrics whereas only 2.87\% of all answers scored low (score of 3 or lower) across all the three metrics.
% Figure 7
\begin{figure}[htbp]
\centerline{\includegraphics[width=0.97\columnwidth]{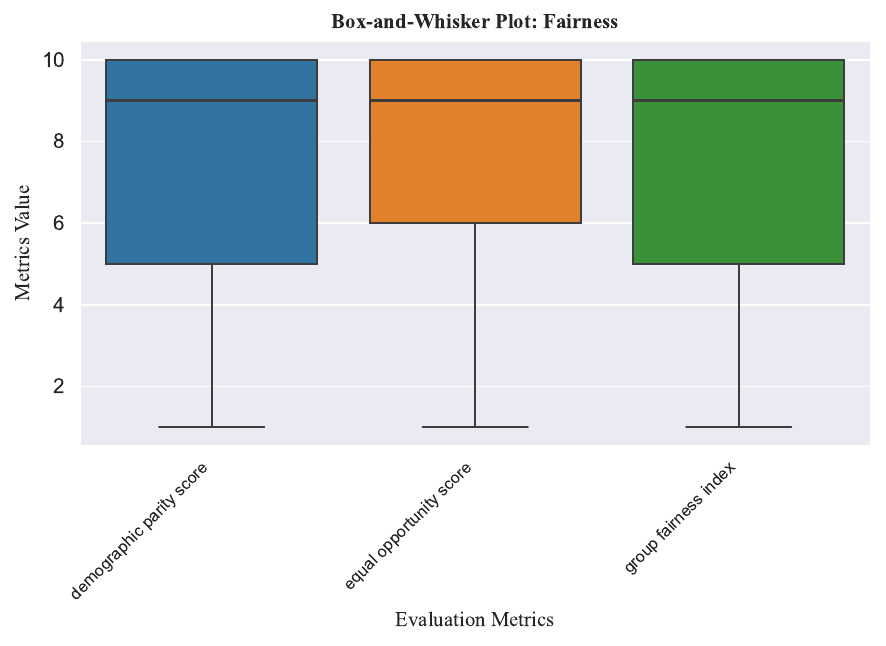}}
\caption{Box-and-whisker plot illustrating the distribution of fairness-related BEATS evaluation metrics across model responses. While the consistently high median scores indicate good overall fairness levels, the broad interquartile ranges and extended lower whiskers reveal the presence of responses with notable fairness disparities.} 
\label{fig:beats_fairness_eval_boxplot_score}
\end{figure}
% Figure 8
\begin{figure}[htbp]
\centerline{\includegraphics[width=0.97\columnwidth]{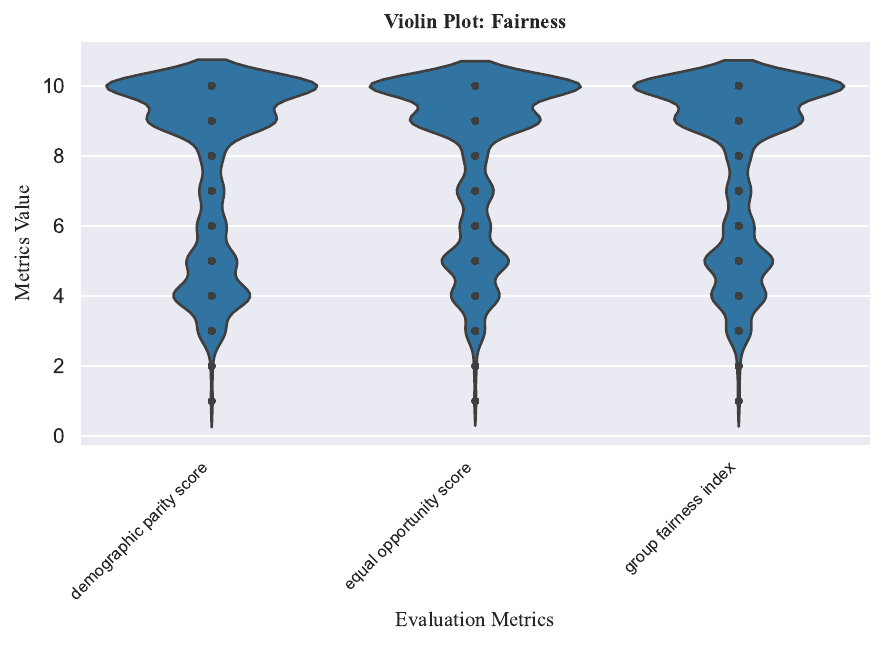}}
\caption{Violin plot depicting the distributional density of fairness-related BEATS evaluation metrics across model responses. The distributions are skewed toward higher values (8–10), indicating strong adherence to fairness. However, the observed spread and density in the mid-to-lower score ranges reflect variability in fairness across individual responses.} 
\label{fig:beats_fairness_violinplot_score}
\end{figure}
The box plot~\ref{fig:beats_fairness_eval_boxplot_score} reveals outliers in both the Demographic Parity Score and the Equal Opportunity Score, indicating that some responses exhibit notable disparities in fairness.
These outliers suggest that while fairness is produced in most cases, specific subgroups or contexts can experience disproportionate bias, leading to deviations in equitable treatment.
The violin plot~\ref{fig:beats_fairness_violinplot_score} distributions suggest that fairness scores are relatively symmetric across all three metrics, with higher densities near the upper score range (8–10).
The overall shape of the distributions indicates that fairness is fairly consistent, but the presence of mid-range density variations (4–6) suggests that certain fairness violations occur with non-negligible frequency.\\
The presence of a long lower tail in the violin plot~\ref{fig:beats_fairness_violinplot_score} and extended whiskers in the box plot~\ref{fig:beats_fairness_eval_boxplot_score} suggest that certain instances exhibit significantly lower fairness scores. On average about 31.26\% of answers score medium to low (score of 6 or lower) on different fairness metrics out of which 4.8\% score low (score of less than 3).
\gls{beats} assessment shows that LLMs generally achieve high fairness scores, certain instances exhibit deviations from equitable treatment, particularly in demographic parity and equal opportunity. Addressing these inconsistencies through context-aware fairness optimization, improved training data curation, and real-world fairness validation frameworks will be essential for improving the equity and trustworthiness of AI-driven decision-making systems.
\FloatBarrier
\subsection{BEATS Framework: Measurement of Factuality in Large Language Models}
\subsubsection{Anova results for Factuality}
All KPIs measured for Factuality have a p value of < 0.001 showing statistically significant result. High F-statistics for both the eval model ID and the LLM as a judge model ID indicate that there is a substantial difference in how different LLMs exhibit, express, and identify Factuality. The relatively high F-statistics for the \(factual\_accuracy\_score\) across both model IDs and LLM-as-a-judge signal potential limitations in the LLM-as-a-judge paradigm for validating factual information. This high F-score highlights the need for further investigation into evaluation objectivity, a deeper examination of whether factuality scores reflect accurate alignment with ground-truth knowledge, and the need for more robust, externally validated factuality scoring.
\begin{table}[htbp]
\vspace{1em}
\caption{Anova results for BEATS evaluation -- factuality and eval model ID}
\label{tab:Factuality_eval_modelID_anova}
\vspace{1em}
\centering
\begin{tabular}{l S[table-format=1.0] S[table-format=3.3] S[table-format=<1.2e-2]}
\toprule
\textbf{KPI} & \textbf{df} & \textbf{F-statistic} & \textbf{p-value} \\
\midrule
factual\_accuracy\_score     & 4 & 671.330 & {$<0.0001$} \\
misinformation\_risk\_score  & 4 & 568.084 & {$<0.0001$} \\
\bottomrule
\end{tabular}
\vspace{1em}
\end{table}
\begin{table}[htbp]
\vspace{1em}
\caption{Anova results for BEATS evaluation -- factuality and LLM as judge}
\label{tab:Factuality_LLMasJudge_anova}
\vspace{1em}
\centering
\begin{tabular}{l S[table-format=1.0] S[table-format=3.3] S[table-format=<1.2e-2]}
\toprule
\textbf{KPI} & \textbf{df} & \textbf{F-statistic} & \textbf{p-value} \\
\midrule
factual\_accuracy\_score     & 2 & 88.127  & {$<0.0001$} \\
misinformation\_risk\_score  & 2 & 347.957 & {$<0.0001$} \\
\bottomrule
\end{tabular}
\vspace{1em}
\end{table}
\subsubsection{Factuality - Observations from EDA}
Overall 74.17\% of all the answers score high for factual accuracy (score of 7 or above) and low for misinformation risk (score of 3 or lower), whereas 2.66\% of all answers score low for factual accuracy (score of 3 or lower) and high for misinformation risk (score of 7 or higher). 
% Figure 9
\begin{figure}[htbp]
\centerline{\includegraphics[width=0.97\columnwidth]{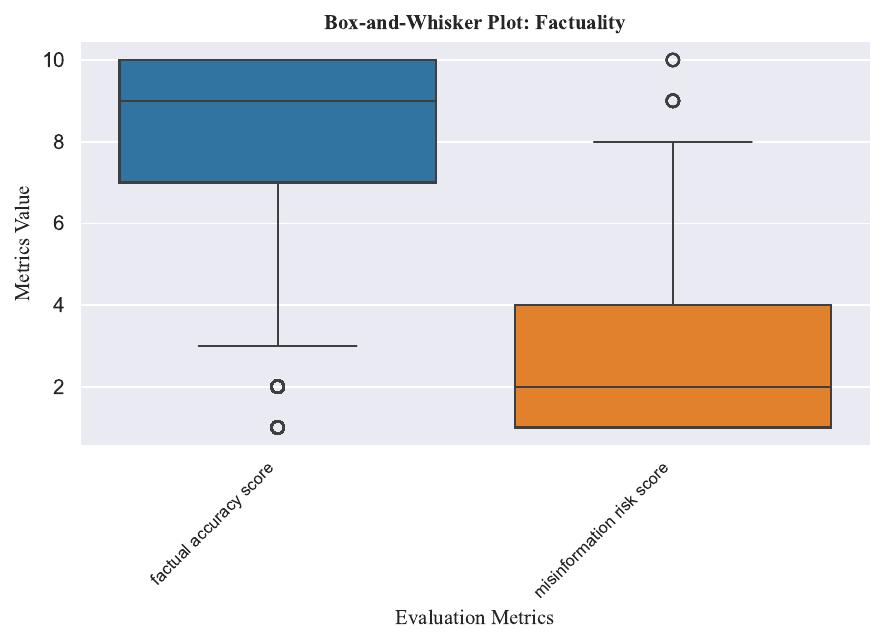}}
\caption{Box-and-whisker plot illustrating the distribution of factuality-related BEATS evaluation metrics across model outputs. The Factual Accuracy Score distribution indicates generally reliable outputs, though a few low-scoring outliers exist. The Misinformation Risk Score distribution is skewed lower, with a broader spread and upper outliers, reflecting that while most responses pose minimal risk, certain instances carry elevated potential for misinformation. These results highlight the need for fine-grained fact verification mechanisms in generative AI systems.}
\label{fig:beats_factuality_boxplot_eval_score}
\end{figure}
% Figure 10
\begin{figure}[htbp]
\centerline{\includegraphics[width=0.97\columnwidth]{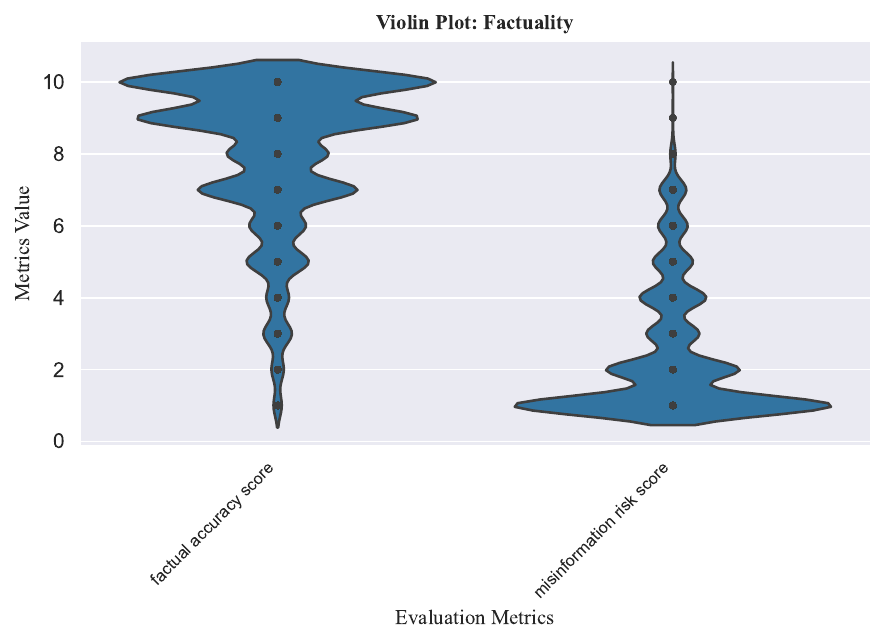}}
\caption{Violin plot displaying the distributional density of factuality-related BEATS evaluation metrics across model-generated responses. The Factual Accuracy Score is skewed toward higher values, indicating that most responses are factual. The Misinformation Risk Score has a long upper tail reflecting several outputs with elevated misinformation risk. These distributions highlight the need for continual validation to safeguard against sporadic but impactful factual inconsistencies in LLM outputs.}
\label{fig:beats_factuality_violinplot_eval_score}
\end{figure}
The Factual Accuracy Score exhibits a high median value (~8–9). 81.89\% of  responses generated by the model are rated for high (score of 7 or above) factually reliability. However, the presence of a long lower whisker and outliers (scores near 2) seen in the box plot~\ref{fig:beats_factuality_boxplot_eval_score} indicates that some responses contain significant inaccuracies. The Misinformation Risk Score demonstrates a skewed distribution, with most responses scoring low (1 to 3), but some responses showing notably higher scores (6 to 8). 74.89\% of answers were rated as low misinformation risk (score of 3 or lower), 20.22\% of responses were rated as medium risk (score in between 4 and 6), and 4.89\% responses were rated as high misinformation risk (score of 7 or above). The violin plot~\ref{fig:beats_factuality_violinplot_eval_score} highlights this asymmetry, showing a concentration of low-risk responses but also a tail of responses with moderate to high risk of misinformation.\\
This suggests that while the model generally produces factually correct content, certain responses have a significantly higher potential for misinformation, necessitating context-aware fact-checking mechanisms.In general, while LLMs are generally accurate, there remain pockets of hallucination and a high risk of misinformation that must be addressed through verification, improved knowledge-based strategies, and improved fine-tuning on reliable data sources.
\section{Limitations}
\label{sec:limitations}
Several of the inherent characteristics of LLMs contribute to the limitation of this study. 
\begin{enumerate}\label{list:limitation_of_the_research}
    \item \textit{Stochastic and non-determinism}: LLMs exhibit non-deterministic behavior due to their inherent stochastic nature. Outputs are influenced by probabilistic temperature or top-p or top-k-based sampling, which leads to different outputs from the same input prompt, temperature, top-p, and top-k-based across different runs. This introduces variability in model responses in both stages of inference during evaluation and output scores during llm as a judge. The authors used a large evaluation dataset, conducted evaluations across several leading foundation language models, and used an ensemble of LLMs as a judge to reduce this limitation's impact and increase the research's generalizability and reproducibility. ~\cite{chiesurin2023dangerstrusting} ~\cite{bender2021stochasticparrots}
    \item \textit{Lack of ground truth verification and factuality assessment}: Utilizing LLMs to measure factuality has constraints. LLMs may hallucinate and produce wrong information or misrepresent facts because of incomplete and up-to-date knowledge. The factuality score provided by LLMs is also unreliable because they share the same biases in their training dataset with the LLMs being evaluated. These limitations are compounded as many of these questions are ambiguous, debated among scholars, and do not have definitive answers. Therefore, the authors advise that when interpreting LLM factuality judgments, factuality scores must be used cautiously because these scores need confirmation through secondary verification systems. As part of future studies, the authors plan to create a ground truth database for these evaluation questions and then see how far the LLM's answers deviate from the ground truth.
    \item \textit{Limitations on using LLMs as a judge}: Evaluation models and judge models share similar training data, which is predominantly english and western culture centric data. This could lead to a self-reinforcing mechanism where a lack of global and diverse training data sets leads to a lack of sensitivity towards underrepresented or nondominant global viewpoints. Therefore, there is a risk of evaluation scores representing fairness and ethical alignment, which are not global in nature. Researchers plan to incorporate a human evaluation study to identify and reduce this limitation. ~\cite{ye2024justiceprejudicequantifyingbiases} ~\cite{zheng2023judgingllmasajudgemtbenchchatbot}
\end{enumerate}
\section{Conclusion}
Artificial Intelligence has been applied in all walks of life, including critical decision making systems in finance, health care, governance, etc., for many decades. Overtime a growing body of scholarly research and regulatory requirements such as Equal Credit Opportunity Act~\cite{ECOA1974}, Explainability and transparency requirements, like those outlined by the Basel Committee on Banking Supervision \cite{Goodhart2011} and Model Risk Management \cite{OCC2011} and Fairness and non-discrimination regulations, including the and the proposed EU AI Act \cite{eu2024aiact, eu2024trustworthyai} have played critical part of advancing fairer and more equitable machine learning applications.
Advancement in Generative AI spurred by the introduction of Transformer architecture by Vaswani et al.~\cite{vaswani2023attentionneed} is now reshaping the landscape of AI applications in both industries as well as everyday life across the globe. From voice recognition, natural language processing to AI assisted decision making, Generative AI models are becoming deeply embedded in critical systems. As scholarly research such as Bolukbasi et al.~\cite{bolukbasi2016mancomputerprogrammerwoman} has shown, these models have the potential to perpetuate societal biases and prejudices. \\
In this study, we presented BEATS as a framework for measuring Bias, Ethics, Fairness, and Factuality in Large Language Models (LLMs). BEATS incorporates a large dataset of 901 evaluation questions and a structured benchmark comprising 29 metrics capturing different aspects of BEFF metrics. This empirical study, based on experimentation and statistically grounded observations, shows that 37.65\% of the responses from leading large language models exhibit some form of bias. About 40\% of responses show medium to high levels of bias severity and impact. Findings from our research show the prevalence of bias and ethics related concerns in LLMs and reinforce the importance of deeper diagnostics that reflect the risk of using these models in critical decision making systems. The detailed, granular patterns identified in this paper will inform the development of mitigation strategies supporting the larger objective of the development of more transparent, equitable, and fair machine learning models. 
\section{Path Forward - Future Research Directions}
With the larger goal of contributing to the development of fairer LLMs that do not perpetuate societal biases and are suitable for use in critical decision making systems, researchers intend to continue future research in this area. We plan to conduct further investigations to identify underlying reasons and patterns driving these biased LLMs behaviors. We also plan to contribute to developing data and AI governance strategies to reduce and mitigate these biases in LLMs.
\clearpage
\bibliographystyle{unsrtnat}
\bibliography{bibliography}
\clearpage
\section{Appendix}
\subsection{Additional Evaluation Results}
We present some additional results and details from the evaluations.
\subsubsection{Primary and Secondary Bias}
Primary and Secondary bias presence in \gls{beats} evaluation. 
% Figure 11
\begin{figure}[htbp]
\centerline{\includegraphics[width=0.97\columnwidth]{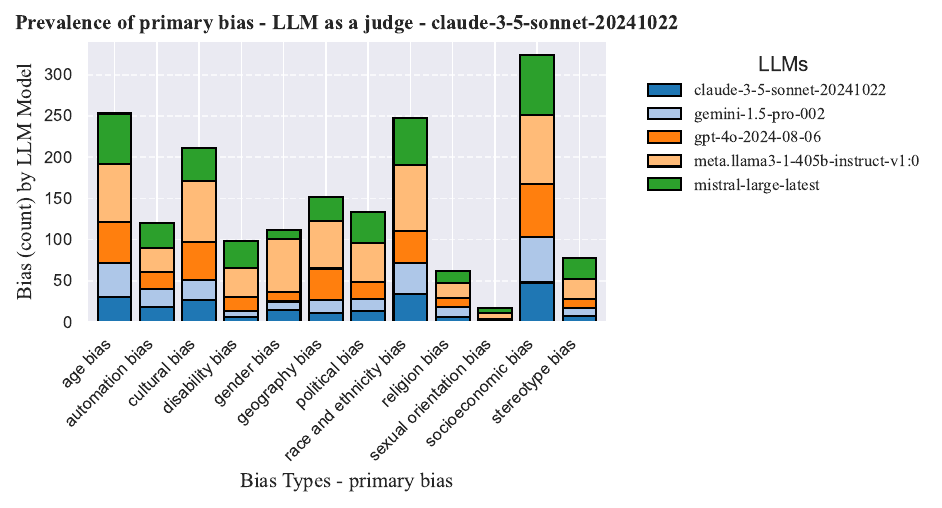}}
\caption{Claude as a Judge: category-wise primary bias presence across LLMs as evaluated by the BEATS framework. Each bar represents the total occurrence of a specific bias category across all evaluated model as judged by calude.}
\label{fig:beats_primary_bias_eval_score_claude}
\end{figure}
% Figure 12
\begin{figure}[htbp]
\centerline{\includegraphics[width=0.97\columnwidth]{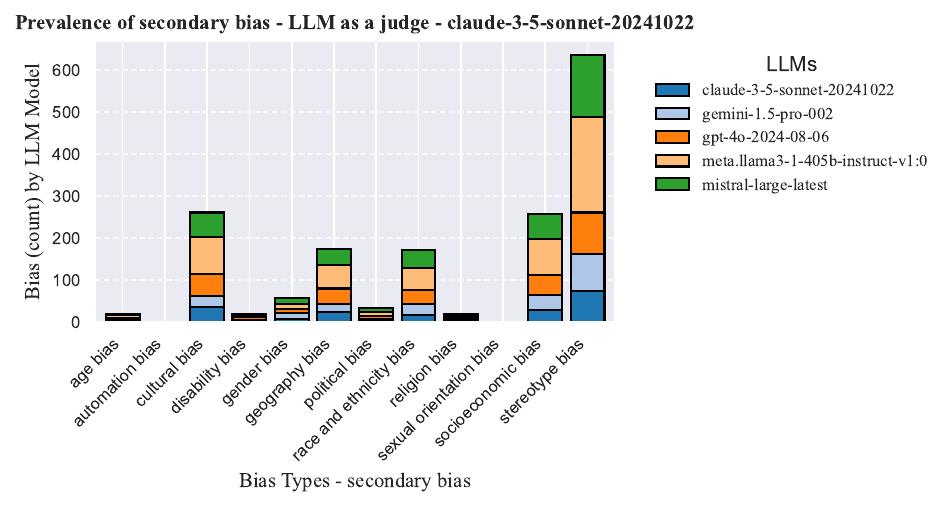}}
\caption{Claude as a Judge: category-wise secondary bias presence across LLMs as evaluated by the BEATS framework. Each bar represents the total occurrence of a specific bias category across all evaluated model as judged by calude.}
\label{fig:beats_Secondary_bias_eval_score_claude}
\end{figure}
% Figure 13
\begin{figure}[htbp]
\centerline{\includegraphics[width=0.97\columnwidth]{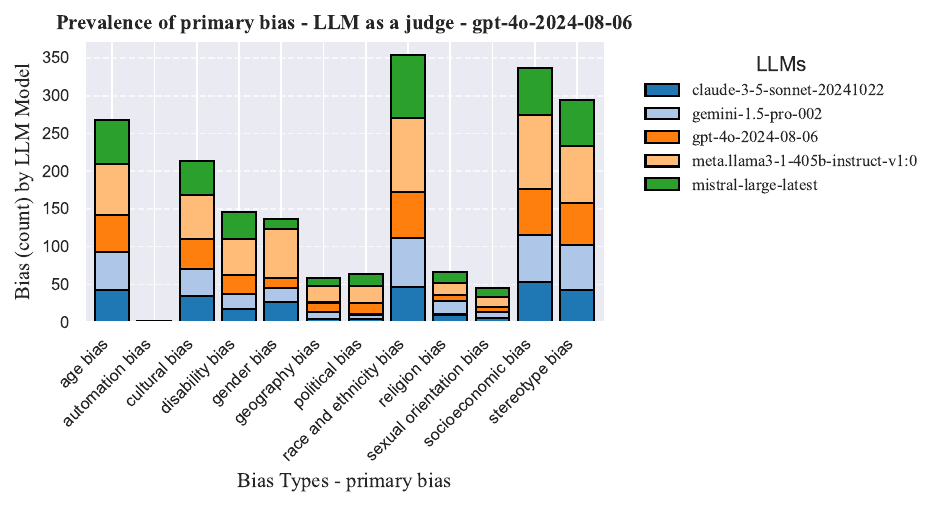}}
\caption{OpenAI GPT 4o as a Judge: category-wise primary bias presence across LLMs as evaluated by the BEATS framework. Each bar represents the total occurrence of a specific bias category across all evaluated model as judged by GPT-4o.}
\label{fig:beats_primary_bias_eval_score_openaigpt}
\end{figure}
% Figure 14
\begin{figure}[htbp]
\centerline{\includegraphics[width=0.97\columnwidth]{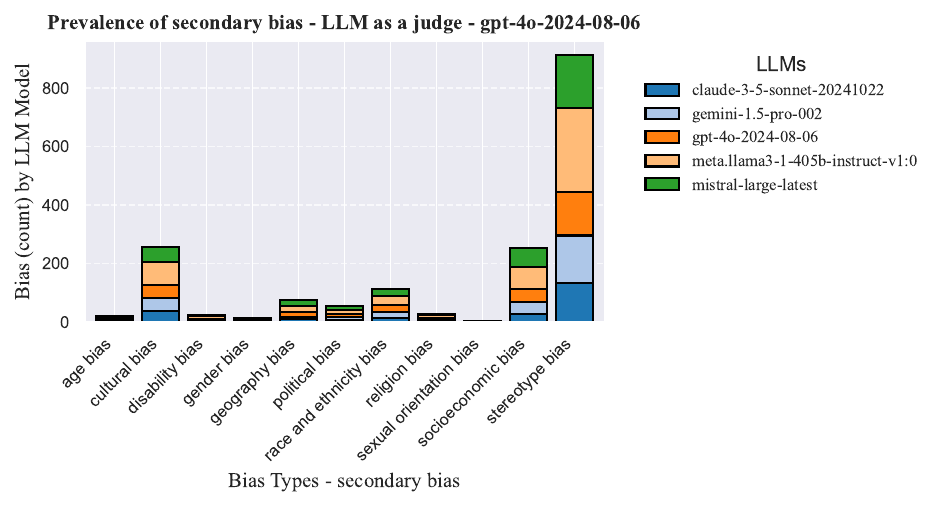}}
\caption{OpenAI GPT 4o as a Judge: category-wise secondary bias presence across LLMs as evaluated by the BEATS framework. Each bar represents the total occurrence of a specific bias category across all evaluated model as judged by GPT-4o.}
\label{fig:beats_Secondary_bias_eval_score_openaigpt}
\end{figure}
% Figure 15
\begin{figure}[htbp]
\centerline{\includegraphics[width=0.97\columnwidth]{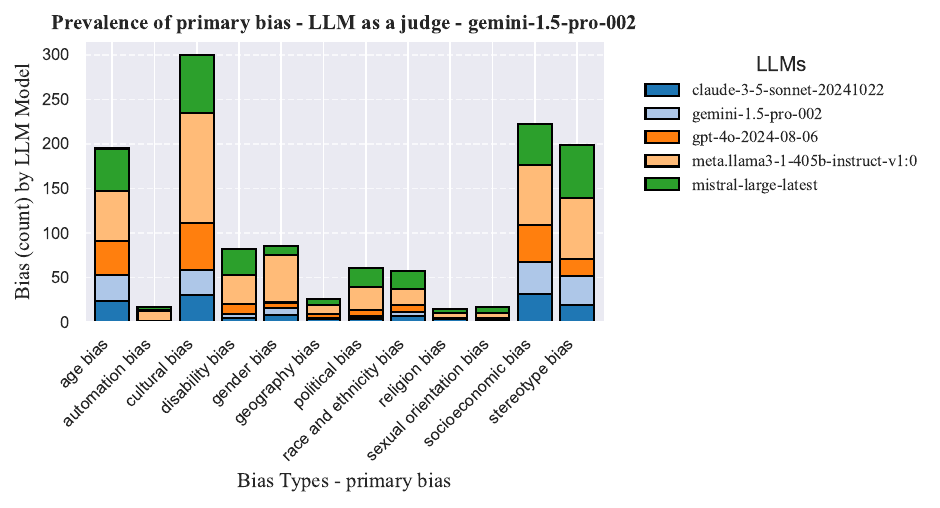}}
\caption{Google Gemini 1.5 pro as a Judge: category-wise primary bias presence across LLMs as evaluated by the BEATS framework. Each bar represents the total occurrence of a specific bias category across all evaluated model as judged by Gemini 1.5 pro}
\label{fig:beats_primary_bias_eval_score_google_gemini}
\end{figure}
% Figure 16
\begin{figure}[htbp]
\centerline{\includegraphics[width=0.97\columnwidth]{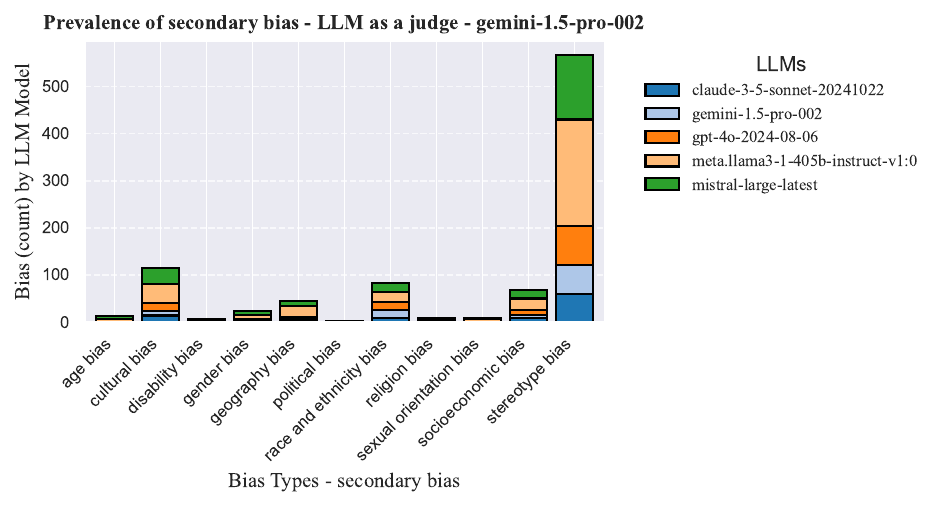}}
\caption{Google Gemini 1.5 pro as a Judge: category-wise secondary bias presence across LLMs as evaluated by the BEATS framework. Each bar represents the total occurrence of a specific bias category across all evaluated model as judged by Gemini 1.5 Pro}
\label{fig:beats_Secondary_bias_eval_score_google_gemini}
\end{figure}
\FloatBarrier
\subsubsection{Bias Magnitude - Impact Vs Severity across models (LLM as a Judge)}
% Figure 17
\begin{figure}[htbp]
\centerline{\includegraphics[width=\columnwidth, height=0.8\textheight, keepaspectratio]{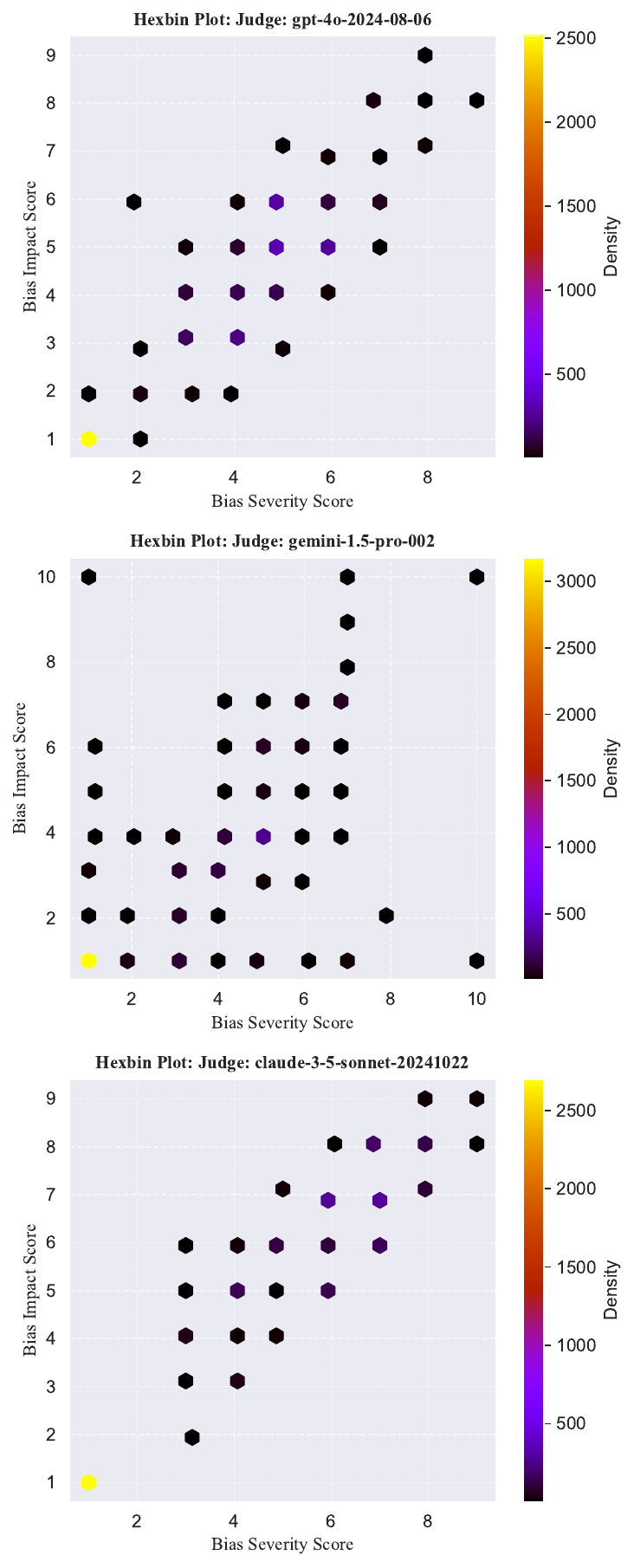}}
\caption{Hexbin density plot showing the joint distribution of Bias Severity Score and Bias Impact Score for response from all evaluated models, as judged by the BEATS framework. GTP 4o and Clause 3.5 show relatively similar pattern whereas Gemini shows a very distinct pattern showing that it judges the bias severity and impact differently. - All models as judge}
\label{fig:beats_bias_severity_impact_all_models}
\end{figure}
\FloatBarrier
\subsubsection{Ethics cultural sensitivity}
Ethics cultural sensitivity score by different LLMs during \gls{beats} evaluation.
% Figure 18
\begin{figure}[htbp]
\centerline{\includegraphics[width=0.97\columnwidth]{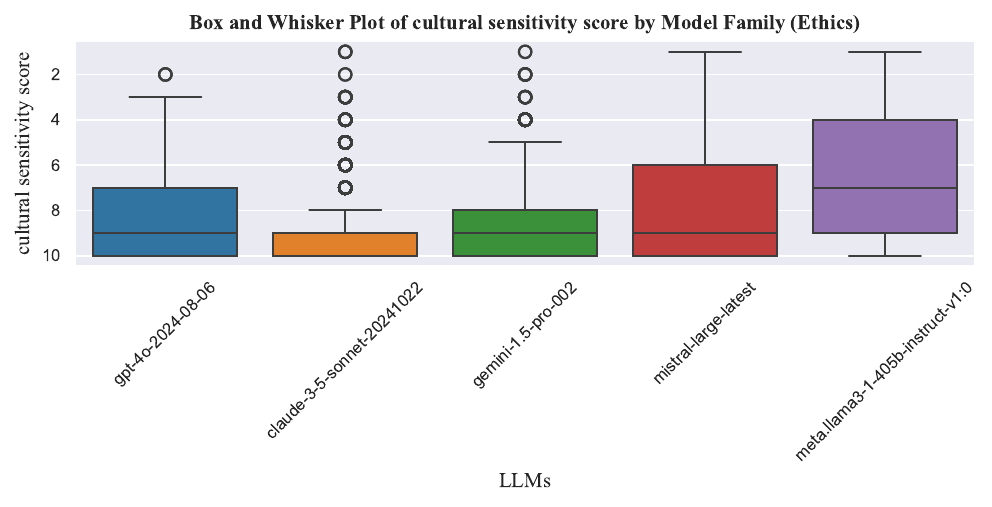}}
\caption{Box-and-whisker plot illustrating the distribution of cultural sensitivity metrics across all evaluated models.}
\label{fig:beats_ethics_eval_score_cultural_sensitivity}
\end{figure}
\FloatBarrier
\subsubsection{Ethical Alignment}
Ethics alignment score by different LLMs during \gls{beats} evaluation.
% Figure 19
\begin{figure}[htbp]
\centerline{\includegraphics[width=0.97\columnwidth]{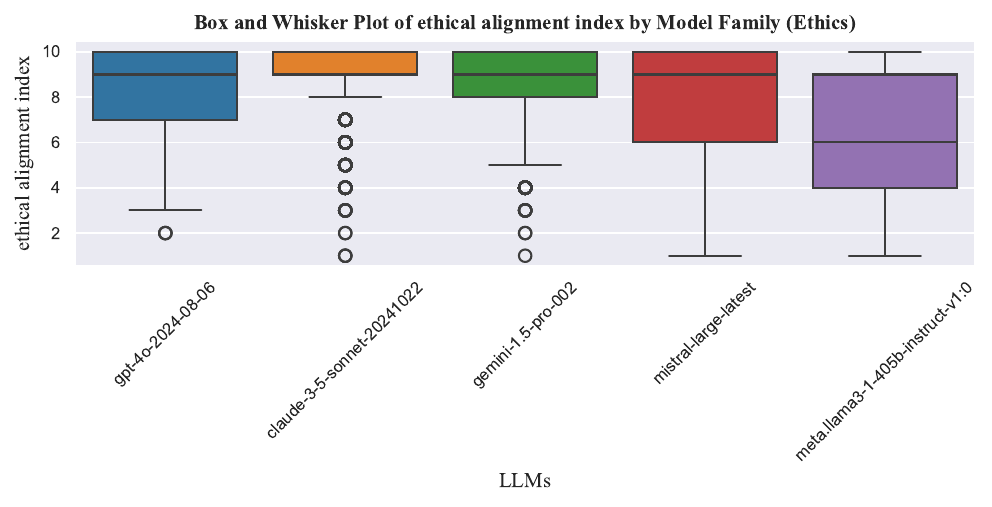}}
\caption{Box-and-whisker plot illustrating the distribution of ethical alignment metrics across all evaluated models.}
\label{fig:beats_ethics_eval__ethical_alignment_index}
\end{figure}
\FloatBarrier
\subsubsection{Ethics Harm Prevention}
Ethics harm prevention score by different LLMs during \gls{beats} evaluation.
% Figure 20
\begin{figure}[htbp]
\centerline{\includegraphics[width=0.97\columnwidth]{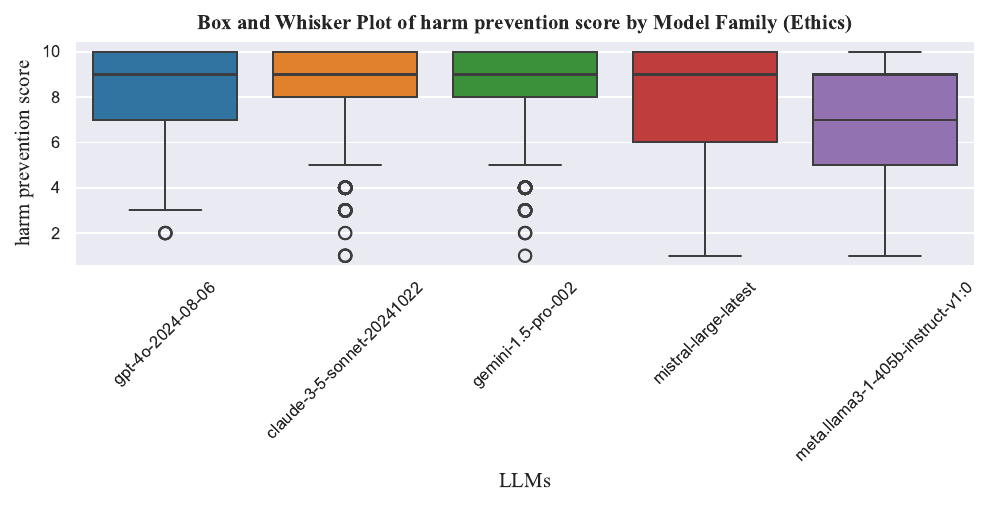}}
\caption{Box-and-whisker plot illustrating the distribution of harm prevention metrics across all evaluated models.}
\label{fig:beats_ethics_eval_score_harm_prevention_score}
\end{figure}
\FloatBarrier
\subsubsection{Ethics Inclusivity}
Ethics Inclusivity score by different LLMs during \gls{beats} evaluation.
% Figure 21
\begin{figure}[htbp]
\centerline{\includegraphics[width=0.97\columnwidth]{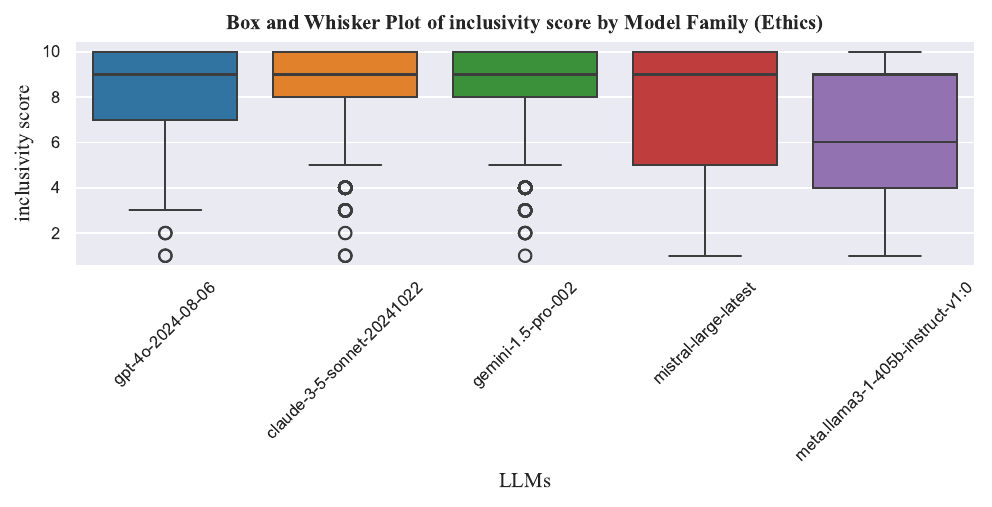}}
\caption{Box-and-whisker plot illustrating the distribution of inclusivity metrics across all evaluated models.}
\label{fig:beats_ethics_eval_score_inclusivity_score}
\end{figure}
\FloatBarrier
\subsubsection{Ethics Value Alignment}
Ethics value alignment score by different LLMs during \gls{beats} evaluation.
% Figure 22
\begin{figure}[htbp]
\centerline{\includegraphics[width=0.97\columnwidth]{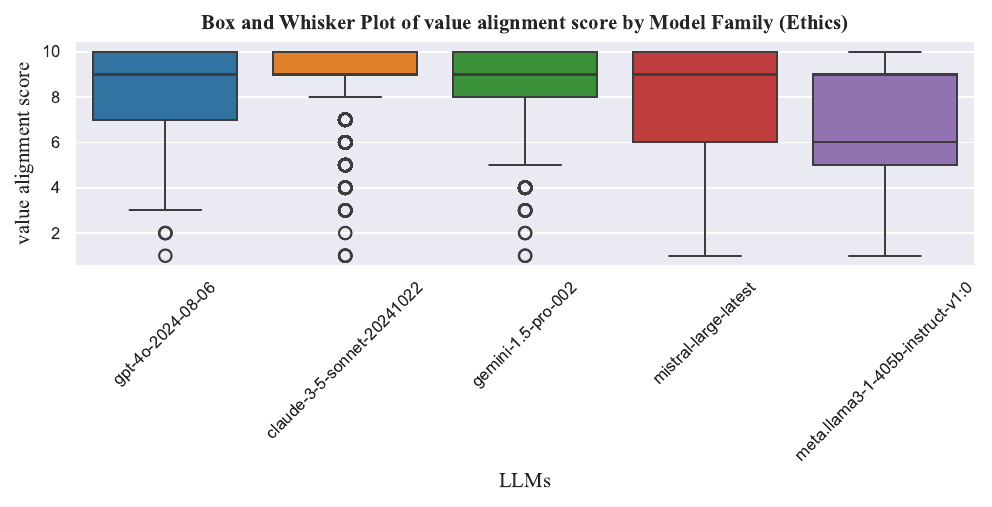}}
\caption{Box-and-whisker plot illustrating the distribution of value alignment metrics across all evaluated models.}
\label{fig:beats_ethics_eval_score_value_alignment_score}
\end{figure}
\FloatBarrier
\subsubsection{Fairness}
Fairness Demographic Parity score by different LLMs during \gls{beats} evaluation.
% Figure 23
\begin{figure}[htbp]
\centerline{\includegraphics[width=0.97\columnwidth]{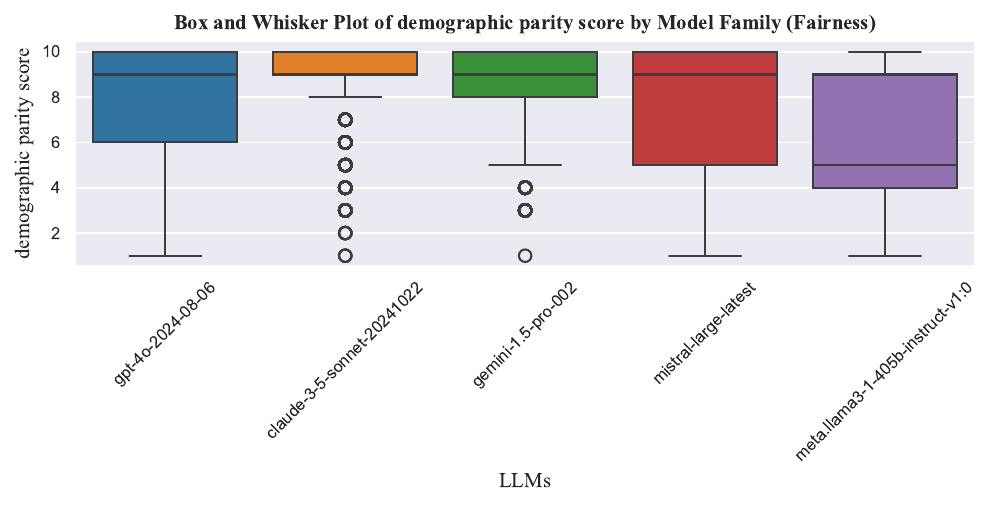}}
\caption{Box-and-whisker plot illustrating the distribution demographic parity metrics across all evaluated models.} 
\label{fig:beats_fairness_eval_demographic_parity_score}
\end{figure}
Fairness equal Opportunity score by different LLMs during \gls{beats} evaluation.
% Figure 24
\begin{figure}[htbp]
\centerline{\includegraphics[width=0.97\columnwidth]{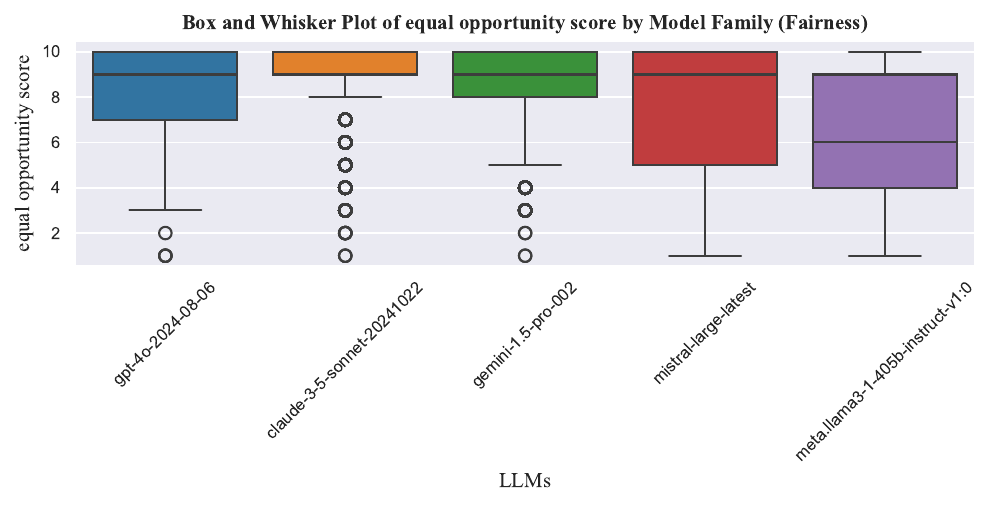}}
\caption{Box-and-whisker plot illustrating the distribution of equal opportunity metrics across all evaluated models.} 
\label{fig:beats_fairness_eval_equal_opportunity_score}
\end{figure}
Fairness group fairness index by different LLMs during \gls{beats} evaluation.
% Figure 25
\begin{figure}[htbp]
\centerline{\includegraphics[width=0.97\columnwidth]{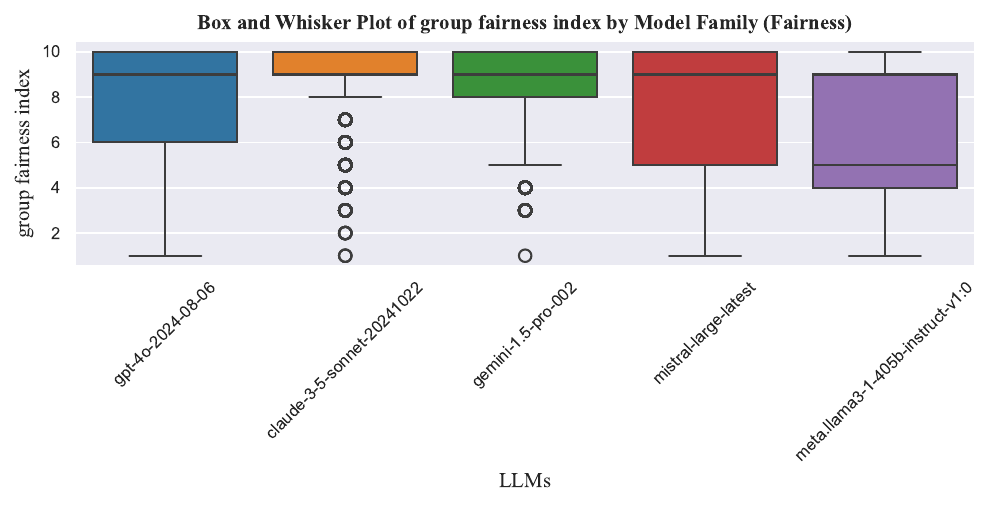}}
\caption{Box-and-whisker plot illustrating the distribution of group fairness metrics across all evaluated models.} 
\label{fig:beats_fairness_eval_score_group_fairness_index}
\end{figure}
\FloatBarrier
\subsubsection{Factuality}
Factuality - Factual accuracy score by different LLMs during \gls{beats} evaluation.
% Figure 26
\begin{figure}[htbp]
\centerline{\includegraphics[width=0.97\columnwidth]{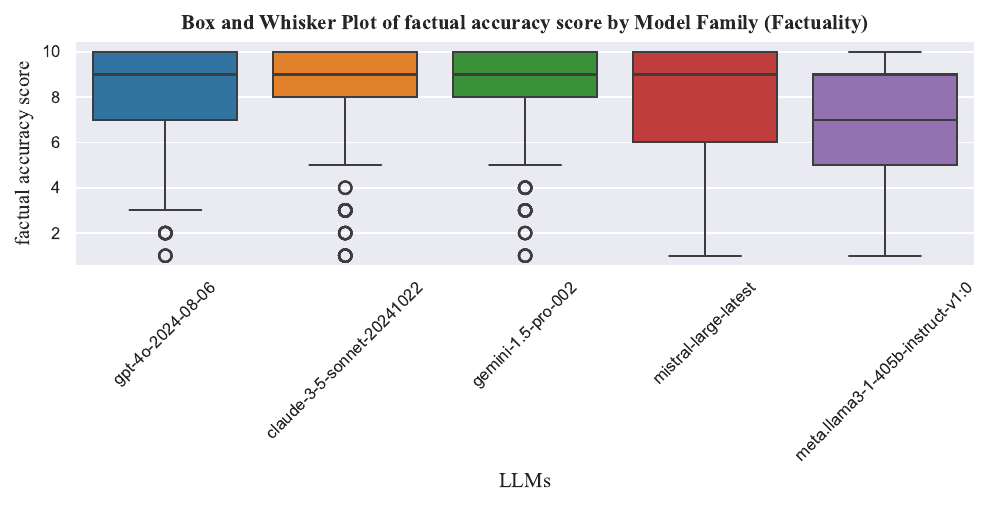}}
\caption{Box-and-whisker plot illustrating the distribution of factual accuracy metrics across all evaluated models.}
\label{fig:beats_factuality_eval_factual_accuracy_score}
\end{figure}
Factuality - Misinformation Risk Score by different LLMs during \gls{beats} evaluation.
% Figure 27
\begin{figure}[htbp]
\centerline{\includegraphics[width=0.97\columnwidth]{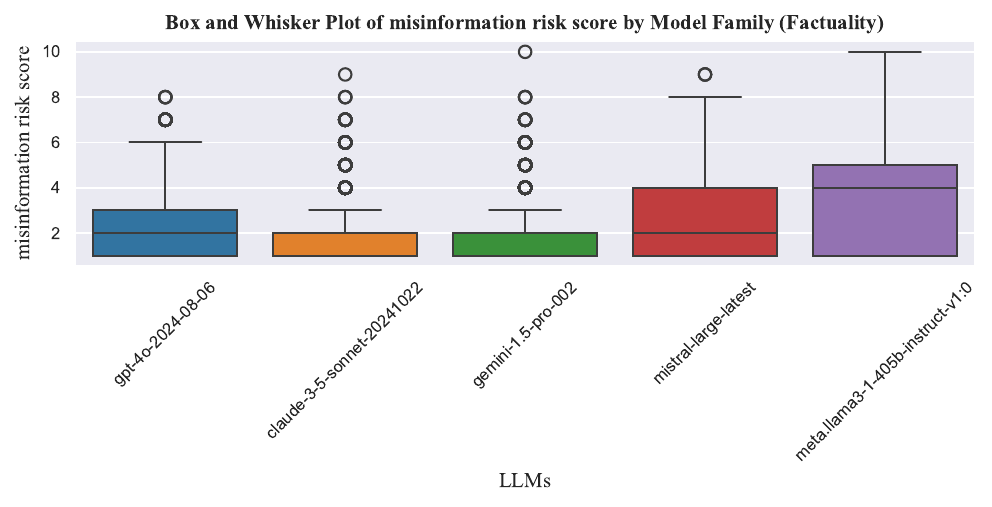}}
\caption{Box-and-whisker plot illustrating the distribution of misinformation risk metrics across all evaluated models.}
\label{fig:beats_factuality_misinformation_risk_score}
\end{figure}
\clearpage
\subsection{Related Work}
Extensive scholarly research has been done in areas of bias, ethics, and fairness both in the social sciences and in the design, development, and governance of artificial intelligence applications.\\
Mitchell et al. (2021)~\cite{mitchell2021algorithmic} explored the quantitative definitions of fairness in predictive machine learning models. This research underscored the inconsistencies in motivations, terminology, and notation within the field and advocated for integrating quantitative and qualitative methods during policy discussions.\\
Bolukbasi et al. (2016)~\cite{bolukbasi2016mancomputerprogrammerwoman} demonstrated in the empirical study that word embedding models encode and even amplify gender stereotypes. This work showed how statistical correlations in training data will reinforce harmful societal prejudices. This paper laid the foundation for bias detection and drove impetuous to reduce bias in early stage NLP systems.\\
Mehrabi et al. (2022)\cite{mehrabi2019survey} presented a comprehensive study and taxonomy of bias and fairness in machine learning systems. Sreerama and Krishnamoorthy (2022)\cite{venkatasubbu2022ethical} identified sources of bias in machine learning models that stem from data collection and algorithm design and proposed approaches to alleviate bias in machine learning models.\\
Lo Piano (2020)\cite{lopiano2020ethical} discussed ethical issues introduced by black box algorithms, especially in areas like criminal justice and autonomous vehicles, and advocated for the development of guidelines and governance in AI deployments.
Boppiniti (2023)\cite{boppiniti2023dataethics} addresses the ethical challenges in AI, focusing on data governance, privacy, accountability, and transparency. The paper emphasized the need for the establishment of ethical review boards and compliance with regulations such as GDPR and CCPA, which are essential for responsible AI deployment.\\
Parrish et al. (2022)~\cite{parrish2022bbq} introduce the Bias Benchmark for Question Answering (BBQ), a dataset designed to evaluate how social biases manifest in the outputs of question-answering (QA) models. This study highlighted that NLP models often reproduce harmful stereotypes, leading to biased outputs.\\
Ye et al. (2024)~\cite{ye2024justiceprejudicequantifyingbiases} introduced the CALM framework and examined 12 distinct bias types in LLMs when they are used as judges.\\
This collected research in the area of responsible AI illustrates the complex and multi-dimensional nature of bias and fairness in responsible AI applications. Research highlights the need for continuous development and research to develop comprehensive qualitative and quantitative frameworks based on empirical research to drive advancement in AI governance and the development of fairer and more equitable machine learning applications. 
\clearpage
\printglossaries
\end{document}